\documentclass[11pt]{article}

\usepackage[preprint]{acl}

\usepackage{times}
\usepackage{latexsym}

\usepackage[T1]{fontenc}

\usepackage[utf8]{inputenc}

\usepackage{microtype}

\usepackage{inconsolata}

\usepackage{graphicx}

\usepackage{stfloats}

\usepackage{placeins}
\usepackage{bibunits}
\defaultbibliography{custom}
\defaultbibliographystyle{acl_natbib}
\usepackage{caption} 
\usepackage{url}
\usepackage{amsmath}    
\usepackage{xcolor}     
\usepackage{algorithm}  
\usepackage{algorithmic} 
\usepackage{amsfonts}
\usepackage{newfloat}
\usepackage{listings}
\usepackage{graphicx}
\usepackage{booktabs}
\usepackage{multirow}
\usepackage[table]{xcolor}
\usepackage{array}
\usepackage{wrapfig}
\usepackage{subfigure}
\usepackage{enumitem}
\usepackage{siunitx}
\usepackage{fontawesome5}

\usepackage{xcolor}
\definecolor{myred}{HTML}{8C1007}

\title{SPARTA: Evaluating \underline{R}easoning \underline{S}egmentation Robustness through Black-Box \underline{A}dversarial \underline{P}araphrasing in \underline{T}ext \underline{A}utoencoder Latent Space}

\author{%
\textbf{Viktoriia Zinkovich}\textsuperscript{$*$} \quad
\textbf{Anton Antonov}\textsuperscript{$*$} \quad
\textbf{Andrei Spiridonov}\textsuperscript{$*$} \quad 
\textbf{Denis Shepelev} \quad
\textbf{Andrey Moskalenko} \\
\textbf{Daria Pugacheva} \quad
\textbf{Elena Tutubalina} \quad
\textbf{Andrey Kuznetsov} \quad
\textbf{Vlad Shakhuro}\textsuperscript{$\dagger$} \\
\\
 \textsuperscript{$*$}Equal contribution\quad
 \textsuperscript{$\dagger$}Project leader
}

\begin{document}
\maketitle
\begin{abstract}
Multimodal large language models (MLLMs) have shown impressive capabilities in vision-language tasks such as \textit{reasoning segmentation}, where models generate segmentation masks based on textual queries. 
While prior work has primarily focused on perturbing image inputs, \textit{semantically equivalent} textual paraphrases—crucial in real-world applications where users express the same intent in varied ways—remain underexplored.
To address this gap, we introduce a novel \emph{adversarial paraphrasing task}: generating grammatically correct paraphrases that preserve the original query meaning while degrading segmentation performance. 
To evaluate the quality of adversarial paraphrases, we develop a comprehensive automatic evaluation protocol validated with human studies.
Furthermore, we introduce \textbf{SPARTA}—a black-box, sentence-level optimization method that operates in the low-dimensional semantic latent space of a text autoencoder, guided by reinforcement learning.
SPARTA achieves significantly higher success rates, outperforming prior methods by up to $\boldsymbol{2\times}$ on both the ReasonSeg and LLMSeg-40k datasets.
We use SPARTA and competitive baselines to assess the robustness of advanced reasoning segmentation models.
We reveal that they remain vulnerable to adversarial paraphrasing—even under strict semantic and grammatical constraints. 
All code and data will be released publicly upon acceptance.
\end{abstract}

\begin{bibunit}

\section{Introduction}
\label{sec:intro}

In recent years, foundation models have achieved significant advances across diverse domains of deep learning.
Advances in image classification~\citep{dosovitskiy2021an, liu2022convnet, woo2023convnext} and interactive segmentation~\citep{Kirillov_2023_ICCV, ravi2024sam}, together with progress in large language models (LLMs)~\citep{brown2020language, touvron2023llama, guo2025deepseek, dubey2024llama}, have paved the way for multimodal large language models (MLLMs)~\citep{liu2023visual, liu2024improved, bai2023qwen, wang2024qwen2, li2023blip, peng2023kosmos, lai2024lisa} that seamlessly integrate vision and language.
These models are now integral to diverse applications, including conversational systems like ChatGPT, autonomous driving~\citep{mu2024most, seff2023motionlm, hwang2024emma}, and robot control~\citep{driess2023palm, brohan2022rt, brohan2023rt, black2024pi_0}.
As these models continue to mature, new tasks are emerging—particularly in robotics—that require sophisticated visual perception and reasoning capabilities.
One such task is \emph{reasoning segmentation}~\citep{lai2024lisa}, where a model outputs a binary segmentation mask driven by an implicit text query that requires intricate logical or contextual interpretation.

\begin{figure}[!t]
  \centering
  \includegraphics[width=\linewidth]{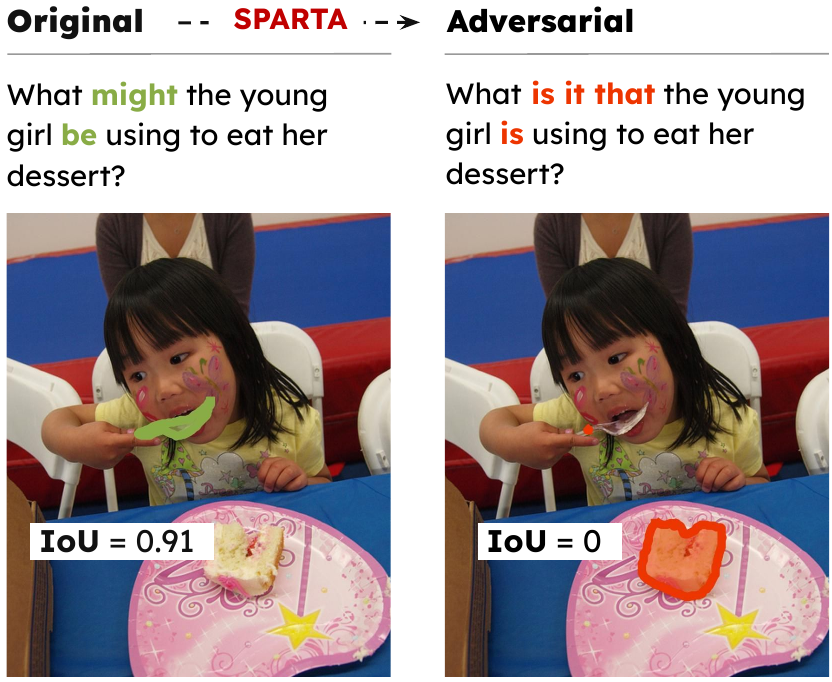}
    \captionsetup{type=figure}
    \vspace{-0.5cm}
  \caption{
  \textbf{Example of an adversarial paraphrase generated by our proposed SPARTA method.}
  The SPARTA produces grammatically correct paraphrases that preserve the original semantic content while significantly degrading segmentation performance.
  }
  \label{fig:overview}
\end{figure}


The quality of a model's predicted segmentation mask is expected to remain consistent, even when users paraphrase their prompts, preserving the original meaning and intent.  
However, the robustness of reasoning segmentation models to query \textit{paraphrasing} remains largely unexplored.


To address this problem, we propose a novel task, \textit{adversarial paraphrasing}, which constrains textual perturbations according to the following criteria:
    (1) the core meaning of the original prompt must be preserved;
    (2) the paraphrase must remain grammatically correct; and
    (3) it must lead to a degradation in segmentation mask predictions.
These constraints enable us to evaluate the robustness of state-of-the-art reasoning-based segmentation models against adversarially paraphrased queries.

Based on this task, we construct a new benchmark to systematically evaluate the robustness of reasoning segmentation models.
We assess state-of-the-art attack strategies, including gradient-based and LLM-based methods; however, these approaches have notable limitations. Gradient-based methods often produce ungrammatical text~\citep{guo-etal-2021-gbda, jones2023arca}, while LLM-based attacks typically rely on heuristic methods, such as iterative refinement~\citep{chao2024jailbreakingblackboxlarge}.


To overcome the limitations of existing gradient-based and heuristic methods, \textbf{we introduce SPARTA}—a novel black-box sentence-level optimization method (Figure~\ref{fig:overview}). 
SPARTA projects queries into a low-dimensional semantic latent space of a pretrained autoencoder and employs reinforcement learning to identify nearby vectors that yield effective adversarial paraphrases.

Overall, our contributions are as follows:
\begin{itemize}
    \item We introduce a novel \textit{adversarial paraphrasing} task designed to evaluate the robustness of reasoning segmentation models against semantically equivalent paraphrased queries.
    To facilitate this evaluation, we introduce an automated evaluation protocol.
    We conduct a user study and demonstrate strong alignment of the proposed scoring method with human judgment.
    
    \item We present a new method for generating adversarial paraphrases, leveraging reinforcement learning-based sentence-level optimization.
    Our method outperforms black-box and white-box baselines by up to $\boldsymbol{2\times}$ on both the \textit{ReasonSeg} and \textit{LLMSeg-40k} datasets, with 2 model‑specific exceptions.
    
    \item We conduct comprehensive experiments to assess the robustness of state-of-the-art reasoning segmentation models under both white-box and black-box adversarial paraphrasing settings.
    Our results indicate that, despite the strict semantic and grammatical constraints, existing reasoning segmentation models remain vulnerable to such attacks.
\end{itemize}


\section{Related Work}

\subsection{Reasoning Segmentation}

In Referring Expression Segmentation (RES), models output segmentation masks from textual descriptions \citep{zou2023xdecoder, rasheed2024glamm, wu2024see, wu2024glee, wang2023hipie, liu2024unilseg}. 
Expanding on RES, the \textit{reasoning segmentation task} was introduced to handle prompts requiring world knowledge and logical reasoning~\citep{lai2024lisa}. 

The pioneering reasoning segmentation model LISA employs an embedding-as-mask paradigm, decoding a \texttt{<SEG>} token via SAM to produce a segmentation mask. 
Several LISA-based models followed, such as LISA++ \citep{yang2024lisa++}, which can incorporate segmentation results into text responses, and GSVA \citep{xia2024gsva}, which introduces a \texttt{<REJ>} token to explicitly reject absent objects.

\subsection{Adversarial Attacks on Text Modality}

Evaluating model robustness often involves adversarial attacks, which are broadly categorized as white-box or black-box, based on the attacker's access to model internals.

White-box attacks leverage gradient information to optimize adversarial paraphrases, addressing the challenges posed by the discrete nature of text through techniques such as Taylor expansion~\citep{ebrahimi-2018-hotflip, jones2023arca} and Gumbel-Softmax sampling~\citep{jang2017categorical, guo-etal-2021-gbda}.
Among these, we consider two state-of-the-art methods: GBDA~\citep{guo-etal-2021-gbda} and ARCA~\citep{jones2023arca}.
While ARCA achieves strong attack success, it lacks semantic regularization, frequently inserting special symbols that undermine its suitability as a paraphrasing baseline.
In contrast, GBDA incorporates semantic similarity constraints; however, it remains limited to token-level substitutions, constraining paraphrase diversity.


In the black-box setting, attacks have progressed from simple word- and character-level manipulations, such as synonym substitution~\citep{jin2020bertreallyrobuststrong, ren-etal-2019-generating} and character edits~\citep{gao2018blackboxgenerationadversarialtext}, to methods that generate semantically equivalent paraphrases using transformer-based models~\citep{li-etal-2020-bert-attack, iyyer-etal-2018-adversarial}.
Recent developments further leverage LLMs to generate more semantically diverse paraphrases~\citep{yan2023parafuzzinterpretabilitydriventechniquedetecting, xu2023llmfoolitself}.
Among these, we consider PAIR~\citep{chao2024jailbreakingblackboxlarge}—a state-of-the-art and widely used adversarial method—and Qwen3-32B~\citep{qwen3technicalreport}, a leading LLM, as attack baselines.
While these techniques often produce fluent paraphrases, they typically depend on heuristic rules or manual trial-and-error, lacking controllable optimization. 
To address this gap, we optimize paraphrases in the low-dimensional semantic latent space of a pretrained text autoencoder, leveraging reinforcement learning to maximize degradation of segmentation performance—thereby enabling more effective adversarial paraphrasing.



\subsection{Evaluation of Adversarial Attacks}

Adversarial attack effectiveness is primarily measured by the attack \textit{success rate} (SR), where success is determined by the model's output quality drop crossing a task-specific threshold. 
For instance, this could be a drop in Intersection over Union (IoU) for interactive segmentation \citep{liu2025regionguidedattacksegmentmodel, Huang2024SegmentSC} or confidence score changes in classification task \citep{guo-etal-2021-gbda, dong2019evading}. 

To ensure the validity and semantic consistency of adversarial paraphrases, we consider text quality metrics.
Semantic preservation is commonly assessed via cosine similarity between embeddings of the original and paraphrased sentences~\citep{guo-etal-2021-gbda, thieu2021lexdivpara, sun2024textualsimilaritykeymetric}.
In practice, many recent studies apply a cosine similarity threshold to filter paraphrases, with the cutoff depending on the embedding model used~\citep{kassem2024findingneedleadversarialhaystack, Herel_2023}. 
Furthermore, recent works leverage LLMs for evaluation through GPT-scoring \citep{fu-etal-2024-gptscore, wang-etal-2023-chatgpt, chiang-lee-2023-large, chan2023chateval}, though the reliability of such metrics remains an open question \citep{wang-etal-2024-large-language-models-fair}. 
To offer a more trustworthy assessment of paraphrase quality, we combine cosine-based filtering with LLM-based scoring into a comprehensive evaluation pipeline.


\section{Proposed Method: SPARTA}
\label{sec:method}

In this section, we introduce \textbf{SPARTA}—a novel black-box paraphrasing method that generates grammatically correct, semantically consistent paraphrases which degrade segmentation performance.
The input query is first encoded into a continuous latent representation using a pretrained text autoencoder (Section~\ref{sec:latent}). 
A set of candidate vectors is sampled from a Gaussian distribution in the latent space, centered at the original latent vector. 
These candidates are decoded into paraphrases and evaluated via a reward function that penalizes overlap with the original segmentation mask, while regularization ensures semantic fidelity.
The policy is optimized via Proximal Policy Optimization to guide the sampling toward more effective adversarial paraphrases (Section~\ref{sec:rl}).
The full optimization pipeline is detailed in Algorithm~\ref{alg:rl-sonar-ppo}.


\subsection{Latent Sentence Space}
\label{sec:latent}

Instead of searching paraphrases over discrete tokens, we operate in the {\em continuous} latent space of a pretrained text autoencoder $(E, D)$. 
The encoder $E$ maps the input query $\mathbf{x}$ to a continuous semantic space, and the decoder $D$ reconstructs the original sentence from the latent vector $\mathbf{z}$:
\begin{equation}
\mathbf{z} = E(\mathbf{x}) \in \mathbb{R}^d,\quad \hat{\mathbf{x}} = D(\mathbf{z}),\quad \hat{\mathbf{x}} \approx \mathbf{x}.
\end{equation}
We adopt SONAR \citep{duquenne2023sonar}, a 1B-parameter multilingual model, as the state-of-the-art text autoencoder $(E, D)$, which is described in detail in Appendix~\ref{app:sonar}. 
Its training objective includes translation and MSE losses on sentence embeddings, encouraging language-agnostic latent representations and a well-aligned cross-lingual embedding space.

For the original input query $\mathbf{x}_0$, we obtain the initial embedding $\mathbf{z}_0 = E(\mathbf{x}_0)$ and optimize the latent vector $\mathbf{z} \in \mathbb{R}^d$, initialized with $\mathbf{z}_0$.

\subsection{Reinforcement Learning Formulation}
\label{sec:rl}

SPARTA learns a stochastic policy \(\pi_{\theta}(\mathbf{z} \mid \mathbf{z}_0)\) that perturbs the original latent vector \(\mathbf{z}_0\) to generate adversarial paraphrases. The policy is modeled as a diagonal Gaussian distribution in latent space:
\begin{equation}
\mathbf{z} \sim \pi_{\theta}(\mathbf{z} \mid \mathbf{z}_0) = \mathcal{N}(\boldsymbol{\mu}, \mathrm{diag}(\boldsymbol{\sigma}^2)),
\end{equation}
where the learnable parameters \(\theta = (\boldsymbol{\mu}, \boldsymbol{\sigma})\) consist of the mean \(\boldsymbol{\mu} \in \mathbb{R}^d\) (initialized with \(\mathbf{z}_0\)) and the standard deviation \(\boldsymbol{\sigma} \in \mathbb{R}^d\).%
\footnote{The scale is reparameterized as \(\boldsymbol{\sigma} = \log(1 + \exp(\boldsymbol{\lambda}))\) to keep it strictly positive, where \(\boldsymbol{\lambda}\in\mathbb{R}^d\) is trainable.}

\paragraph{Reward} 
For each sampled vector $\mathbf{z}$, we generate a candidate paraphrase $\hat{\mathbf{x}} = D(\mathbf{z})$ and pass it through the attacked reasoning segmentation model $f$.
The model outputs a segmentation mask $\hat{\mathbf{m}}$, and the effectiveness of the adversarial paraphrase is quantified by the following \emph{reward}:

\begin{equation}
\label{eq:reward}
 R = -\text{IoU}(\hat{\mathbf{m}}, \mathbf{m}),
\end{equation}

\noindent where $\mathbf{m}$ is the ground truth mask. Higher rewards correspond to lower Intersection-over-Union, thereby encouraging paraphrases that most effectively degrade model performance.

\begin{algorithm}[t]
\caption{PPO in Latent Sentence Space}
\label{alg:rl-sonar-ppo}
\begin{algorithmic}[1]
\REQUIRE Query \(\mathbf{x}_0\), image \(\mathbf{I}\), ground-truth mask \(\mathbf{m}\); autoencoder \((E,D)\); model \(f\); sample size $n$; iteration number $N$; hyperparams \((\epsilon,\lambda_{\text{sim}},\lambda_{\mathrm{adv}},\boldsymbol{\sigma},V_\psi)\)
\STATE Initialize \(\mathbf{z}_0 \gets E(\mathbf{x}_0),\;\boldsymbol{\mu}\gets\mathbf{z}_0\)
\FOR{$t = 1$ to $N$}
    \STATE Sample $\mathbf{z}_i \sim \mathcal{N}(\boldsymbol{\mu}, \mathrm{diag}(\boldsymbol{\sigma}^2))$ for $i = 1..n$ 
    \FOR{$i = 1$ to $n$}
        \STATE Decode \(\hat{\mathbf{x}}_i\gets D(\mathbf{z}_i)\), predict \(\hat{\mathbf{m}}_i\gets f(\mathbf{I},\hat{\mathbf{x}}_i)\)
        \STATE Compute \(R_i\), \(A_i\), \(\rho_i\),  \(l_i\) \textcolor{gray}{(Eqs.~\ref{eq:reward}–\ref{eq:ppo})}
    \ENDFOR
    \STATE $\mathcal{L}_{\mathrm{policy}}\!\gets\!-\lambda_{\mathrm{adv}}\frac{1}{n} \sum \ell_i$; \; $\mathcal{L}_{\mathrm{value}}\!\gets\!\frac{1}{n} \sum (R_i - V_\psi)^2$; \; $\mathcal{L}_{\mathrm{sim}} \gets \lambda_{\mathrm{sim}} \|\boldsymbol{\mu} - \mathbf{z}_0\|^2$
    \STATE Update $\boldsymbol{\mu}, \boldsymbol{\sigma}, \psi$ via Adam on $\mathcal{L}_{\mathrm{final}}$ \textcolor{gray}{(Eq.~\ref{eq:total})}
    \STATE Update old policy: $\pi_{\text{old}} \gets \pi$
    \STATE Save \(\hat{\mathbf{x}}\gets D(\boldsymbol{\mu})\) 
\ENDFOR
\end{algorithmic}
\end{algorithm}

\paragraph{Baseline and Advantage}  
To reduce variance in gradient estimates, we learn a scalar value network \(V_\psi\) as a \emph{baseline} \citep{sutton2018rl}. 
The \emph{advantage} \(A\) is then the normalized difference between the observed reward \(R\) and the baseline \(V_\psi(\mathbf{z})\):

\begin{equation}
\label{eq:advantage}
A = \frac{R - V_\psi \;-\;\mathbb{E}[R - V_\psi]}
             {\mathrm{Std}[R - V_\psi]+\varepsilon}.
\end{equation}

\paragraph{Optimization via PPO}  
To train the latent‐space policy, we employ the standard \emph{clipped surrogate objective} of Proximal Policy Optimization (PPO) \citep{schulman2017ppo, huang2023ppo_clip_optimality}. 
At each update, we sample a batch of \(n\) candidate embeddings \(\{\mathbf{z}_i\}\) from the old policy \(\pi_{\theta_{\text{old}}}\), decode each into a paraphrase, and evaluate its adversarial reward \(R_i\).
For each sample, we compute the importance weight: 

\begin{equation}
\label{eq:imp_weight}
\rho_i = \frac{\pi_\theta(\mathbf{z}_i)}{\pi_{\theta_{\text{old}}}(\mathbf{z}_i)} 
= \exp\bigl(\log\pi_\theta(\mathbf{z}_i) - \log\pi_{\theta_{\text{old}}}(\mathbf{z}_i)\bigr)\!,
\end{equation}  

\noindent then form the clipped surrogate:  

\begin{equation}
\label{eq:ppo}
l_i = \min\!\bigl(\rho_i\,A_i,\;\mathrm{clip}(\rho_i,1-\epsilon,1+\epsilon)\,A_i\bigr).
\end{equation}

\noindent Here, \(\epsilon = 0.2\) is the \emph{clip ratio} hyperparameter, which constrains the policy update to a trust region \([1-\epsilon,1+\epsilon]\). 
Clipping \(\rho_i\) prevents large updates that could destabilize training \ \citep{schulman2015trpo, schulman2017ppo}.

\paragraph{Objective function}  
The final optimization objective $\mathcal{L}_{\mathrm{final}}$ combines three terms:
\begin{equation}  
\label{eq:total}
-\underbrace{\lambda_{\mathrm{adv}}\frac{1}{n}\sum_{i=1}^n l_i}_{\mathcal{L}_{\mathrm{policy}}}
+ \underbrace{\frac{1}{n}\sum_{i=1}^n (R_i - V_\psi)^2}_{\mathcal{L}_{\mathrm{value}}}
+ \underbrace{\lambda_{\mathrm{sim}}\|\boldsymbol{\mu} - \mathbf{z}_0\|_2^2}_{\mathcal{L}_{\mathrm{sim}}},
\end{equation}
where \(\mathcal{L}_{\mathrm{value}}\) trains the baseline and \(\mathcal{L}_{\mathrm{sim}}\) preserves semantic fidelity to the original query. 
Optimization is performed with Adam using separate learning rates for \(\boldsymbol{\mu}\), \(\boldsymbol{\sigma}\), and \(\psi\).

\section{Proposed Evaluation Protocol}
\label{sec:eval_protocol}


In this section, we introduce an automatic evaluation protocol for our novel adversarial paraphrasing task. 
We begin by outlining the main steps of the protocol (Section~\ref{sec:eval:protocol}).
We then examine the challenges associated with its core component, LLM-based paraphrase detection, and propose additional filtering steps to enhance performance (Section~\ref{sec:eval:llm}).
Finally, we evaluate the detection methods and ablate the proposed improvements through human studies (Section~\ref{sec:eval:hs}).

\subsection{Evaluation Protocol}
\label{sec:eval:protocol}





We introduce an \textit{automatic evaluation protocol for the adversarial paraphrasing task}.
Since existing attack methods may produce invalid outputs, we select the best adversarial prompts---based on attack loss and paraphrasing quality---and use them to evaluate attack performance.
Specifically, given a set of adversarial prompts obtained through an attack over N iterations, we proceed as follows:
(1) remove duplicate prompts;
(2) discard any prompt that does not reduce the segmentation model’s IoU;
(3) detect which prompts are valid paraphrases;
(4) select the paraphrase that yields the greatest relative IoU drop.


A critical step in this evaluation protocol is \textit{paraphrase detection}, which, as we demonstrate in the following section, presents significant challenges.

\begin{table*}[!htbp]
\small
\centering
\begin{tabular}{@{}p{\dimexpr .255\linewidth - \tabcolsep\relax}%
                p{\dimexpr .745\linewidth - \tabcolsep\relax}@{}}
\toprule
\textbf{Type} & \textbf{Text} \\
\midrule[0.5pt]
\midrule[0.5pt]

Original & the youngest person \\
\addlinespace[2pt]
\textbf{PAIR paraphrase} & \textcolor{myred}{considering standard human growth patterns, pinpoint the individual who, if all people in the image were lined up in order of birth, would be positioned closest to the beginning of the sequence} \\

\midrule[0.5pt]

Original & the sauce \\
\addlinespace[2pt]
\textbf{PAIR paraphrase} &  \textcolor{myred}{the component that is neither the main ingredient nor the garnish, but is distributed throughout the plate in a somewhat fluid form}
 \\
\bottomrule
\end{tabular}
\caption{
\textbf{Examples of PAIR-generated paraphrases that are overly verbose or abstract.} 
The first paraphrase employs indirect and wordy language, while the second describes the sauce ambiguously without explicitly naming it, leaving it unclear whether it refers to sauce, oil, or dressing. 
Despite this, LLMs rate such paraphrases as valid.
}
\label{tab:llm_paraphrase_examples}
\end{table*}

\begin{table*}[!htbp] 
\centering
\resizebox{\linewidth}{!}{%
\begin{tabular}{@{}l ccc ccc ccc ccc@{}}
\toprule
\multirow{3}{*}{\textbf{Prompt}} 
& \multicolumn{6}{c}{\textbf{Qwen3}} 
& \multicolumn{6}{c}{\textbf{LLaMA-3.1-Nemotron}} \\

\cmidrule(lr){2-7}\cmidrule(lr){8-13}

& \multicolumn{3}{c}{LLM} 
& \multicolumn{3}{c}{LLM \& RegExp \& CosSim} 
& \multicolumn{3}{c}{LLM} 
& \multicolumn{3}{c}{LLM \& RegExp \& CosSim} \\

\cmidrule(lr){2-4}\cmidrule(lr){5-7}\cmidrule(lr){8-10}\cmidrule(lr){11-13}

& Precision & Recall & F-score
& Precision & Recall & F-score
& Precision & Recall & F-score
& Precision & Recall & F-score \\ 
\midrule
1 & 0.480 & 0.964 & 0.641 
  & 0.623 & 0.865 & 0.725 
  & 0.634 & 0.640 & 0.637 
  & 0.798 & 0.604 & 0.687 \\

2 & 0.480 & 0.991 & 0.647 
  & 0.601 & 0.883 & 0.715 
  & 0.520 & 0.712 & 0.601 
  & 0.664 & 0.640 & 0.651 \\

3 & 0.530 & 0.946 & 0.680 
  & 0.671 & 0.847 & \textbf{0.749} 
  & 0.552 & 0.712 & 0.622 
  & 0.726 & 0.622 & 0.670 \\
\bottomrule
\end{tabular}
}
\vspace{1mm}
\caption{
\textbf{Evaluation of paraphrase detection methods.}
We compare LLM-based detection using the baseline system prompt 1 from~\citet{michail-etal-2025-paraphrasus} against our proposed enhanced system prompts (2 and 3), as well with additional filtering based on regular expressions and semantic cosine similarity. Best F-scores are shown in \textbf{bold}.
}
\label{tab:results:prompts}
\end{table*}
\subsection{LLM-based Paraphrase Detection Issues}
\label{sec:eval:llm}


Paraphrasing involves rephrasing a sentence while preserving its original meaning, intent, and grammatical correctness in a clear and concise manner. 
However, automatically assessing whether generated prompts meet these criteria remains a non-trivial task. 
To address this, we explored a state-of-the-art LLM-based evaluation approach, following prior work~\citet{michail-etal-2025-paraphrasus}. 
Our initial experiments revealed three key issues:

\begin{enumerate}
\item \label{llm:issue:amb} 
Defining a \textit{valid} paraphrase for an LLM is challenging, as commonly accepted definitions like ``alternative expressions of the same meaning''~\citep{xu-etal-2015-semeval} are too broad for reliable automated evaluation.
\item \label{llm:issue:cap} 
LLMs often fail to capture differences in capitalization and terminal punctuation (e.g., ``a person is calling someone'' vs. ``A person is calling someone.'').
Because ReasonSeg dataset contain prompts that may be either fragments or complete sentences, we consider an adversarial prompt to be a valid paraphrase only if it preserves both capitalization and terminal punctuation.


\item \label{llm:issue:riddle}
We observe that some paraphrases become excessively long or abstract, occasionally resembling riddles or puzzles, which LLMs often still judge as valid (Table~\ref{tab:llm_paraphrase_examples}).
Although such paraphrases may retain partial semantic overlap with the original, they obscure the intended meaning and hinder clarity, and thus should not be regarded as valid.
\end{enumerate}

Our findings are consistent with the recent work~\citet{michail-etal-2025-paraphrasus}, which demonstrated that even modern LLMs and specialized classification models struggle with the paraphrasing classification task.




    

We address the issues mentioned above as follows:
\begin{enumerate}
    \item To mitigate Issue~\ref{llm:issue:amb}, we \textit{improve the system prompt} used by the LLM. 
    We consider three different system prompts.
    Prompt 1 is a simple zero-shot binary classification prompt, which performed best in prior work~\citet{michail-etal-2025-paraphrasus}.
    Prompts 2 and 3 provide detailed task instructions, a 5-point scoring scale, and 10 in-context examples.
    In the latter two settings, we consider an adversarial prompt a valid paraphrase only if it receives an LLM score of 5.
    The full prompt templates are included in Appendix~\ref{sup:system_prompts}.

    \item To resolve Issue~\ref{llm:issue:cap}, we apply a \textit{regular expression–based filtering} to discard paraphrases that alter capitalization or terminal punctuation.

    \item  
    We mitigate Issue~\ref{llm:issue:riddle} by filtering out semantically distant paraphrases.
    Specifically, we use Qwen3-Embedding-8B~\citep{qwen3embedding}, a state-of-the-art open-source sentence embedding model, to compute semantic similarity.
    Through empirical analysis, we identify an optimal \textit{cosine similarity} threshold of 0.825 (see Appendix~\ref{sup:thres_val} for details).
    This step improves detection performance by removing overly abstract or indirect prompts.
\end{enumerate}
We evaluated LLM-based detection and ablated our improvements with human studies.


\subsection{Ablation}
\label{sec:eval:hs}

We sampled a dataset of 310 pairs of original and adversarial prompts, generated by the proposed SPARTA and baseline methods (see Section~\ref{subsec:training-details}), and manually annotated them for paraphrase validity.
For LLM-based detection, we evaluated two state-of-the-art models: LLaMA-3.1-Nemotron-70B~\citep{wang2024helpsteer2preferencecomplementingratingspreferences} and Qwen3-32B~\citep{qwen3technicalreport}.
For each LLM, system prompt, and filtering configuration (with or without regular expressions and cosine similarity), we measured performance using the F1-score to identify the most effective detection setup. 

The results of the human study are summarized in Table~\ref{tab:results:prompts}.
The best detection performance was achieved using Qwen3-32B with system prompt 3 and filtering based on regular expressions and cosine similarity, yielding an F1 score of 0.749.
Using system prompt 3 without additional filtering already led to a notable improvement over system prompt 1 (F1 score: 0.641 $\xrightarrow{}$ 0.680), and further gains were achieved with the full filtering setup (0.680 $\xrightarrow{}$ 0.749).
This best-performing configuration was adopted in the automatic evaluation protocol described earlier.

\begin{table*}[t!]
\centering
\setlength{\tabcolsep}{1.2mm}
\scriptsize
\resizebox{\textwidth}{!}{%
\begin{tabular}{l *{12}{S[table-format=2.1]}}
\toprule
\multirow{2}{*}{Attacked model}
& \multicolumn{3}{c}{\textbf{GBDA}}
& \multicolumn{3}{c}{\textbf{Qwen3 \textit{(simple)}}}
& \multicolumn{3}{c}{\textbf{Qwen3 PAIR}}
& \multicolumn{3}{c}{\textbf{SPARTA \textit{(ours)}}}
\\
\cmidrule(lr){2-4}
\cmidrule(lr){5-7}
\cmidrule(lr){8-10}
\cmidrule(lr){11-13}
& {mSR} & {SR$_5$} & {SR$_{10}$}
& {mSR} & {SR$_5$} & {SR$_{10}$}
& {mSR} & {SR$_5$} & {SR$_{10}$}
& {mSR $\uparrow$} & {SR$_5$ $\uparrow$} & {SR$_{10}$ $\uparrow$} \\
\midrule
LISA [7B]
& 3.2  & 11.0 & 8.4
& \underline{13.5} & \underline{30.9} & \underline{25.1}
& 13.0 & 25.7 & 22.5
& \textbf{26.6} & \textbf{48.7} & \textbf{42.4} \\

LISA-exp. [7B]
& 3.1  & 10.0 & 7.5
& 11.0 & 25.0 & 21.0
& \underline{16.1} & \underline{32.5} & \underline{25.5}
& \textbf{24.6} & \textbf{49.5} & \textbf{42.5} \\

LISA [13B]
& 2.6  & 4.5  & 4.5
& 9.2  & 24.1 & 18.8
& \underline{11.0} & \underline{26.8} & \underline{21.0}
& \textbf{23.2} & \textbf{46.0} & \textbf{38.4} \\

LISA-exp. [13B]
& 2.9  & 7.6  & 6.2
& 8.7  & 24.6 & 18.5
& \underline{11.4} & \underline{27.5} & \underline{21.8}
& \textbf{25.0} & \textbf{47.4} & \textbf{40.8} \\

LISA++ [7B]
& 0.9  & 2.1  & 1.7
& \underline{10.7} & \underline{20.9} & \underline{19.7}
& 8.8 & 17.4 & 14.0
& \textbf{16.2} & \textbf{29.1} & \textbf{23.9} \\

GSVA [13B]
& 2.2  & 6.4  & 4.6
& 15.6 & 28.3 & 23.1
& \underline{16.0} & \underline{31.8} & \underline{29.5}
& \textbf{27.9} & \textbf{53.2} & \textbf{44.5} \\
\bottomrule
\end{tabular}
}
\vspace{2mm}
\caption{
\textbf{Evaluation results of baselines and the proposed SPARTA on state-of-the-art reasoning segmentation models on the LLMSeg-40k dataset.} 
mSR refers to the area under the curve of the success rate (SR) versus the IoU-drop threshold, computed for adversarial paraphrases with an LLM score of 5. 
SR$_{5}$ and SR$_{10}$ represent the success rate for IoU drops greater than 5\% and 10\%, respectively.
Higher values indicate stronger attacks.
The best results are in \textbf{bold}, the second best are \underline{underlined}.
}
\label{tab:llmseg}
\end{table*}
\begin{table*}[t!]
\centering
\setlength{\tabcolsep}{1.2mm}
\scriptsize
\resizebox{\linewidth}{!}{%
\begin{tabular}{l *{12}{S[table-format=2.1]}}
\toprule
\multirow{2}{*}{Attacked model}
& \multicolumn{3}{c}{\textbf{GBDA}}
& \multicolumn{3}{c}{\textbf{Qwen3 \textit{(simple)}}}
& \multicolumn{3}{c}{\textbf{Qwen3 PAIR}}
& \multicolumn{3}{c}{\textbf{SPARTA \textit{(ours)}}}
\\
\cmidrule(lr){2-4}
\cmidrule(lr){5-7}
\cmidrule(lr){8-10}
\cmidrule(lr){11-13}
& {mSR} & {SR$_5$} & {SR$_{10}$}
& {mSR} & {SR$_5$} & {SR$_{10}$}
& {mSR} & {SR$_5$} & {SR$_{10}$}
& {mSR $\uparrow$} & {SR$_5$ $\uparrow$} & {SR$_{10}$ $\uparrow$} \\
\midrule
LISA [7B]
& 6.8  & 18.7 & 12.6
& 11.2 & \underline{32.9} & \underline{24.9}
& \underline{14.6} & 27.6 & 23.4
& \textbf{25.8} & \textbf{47.4} & \textbf{40.8} \\

LISA-exp. [7B]
& 2.8  & 11.2 & 5.8
& 12.5 & \underline{29.0} & \underline{24.1}
& \textbf{14.6} & 26.8 & 23.2
& \underline{14.0} & \textbf{32.3} & \textbf{26.9} \\

LISA [13B]
& 1.4  & 5.7  & 4.9
& 9.6  & 21.5 & 17.0
& \underline{13.3} & \underline{25.1} & \underline{23.5}
& \textbf{16.3} & \textbf{42.7} & \textbf{33.3} \\

LISA‑exp. [13B]
& 3.4  & 8.8  & 6.7
& 7.6  & \underline{25.4} & 18.8
& \underline{11.3}  & 25.0 & \underline{19.9}
& \textbf{17.5} & \textbf{39.9} & \textbf{32.8} \\

LISA++ [7B]
& 2.5  & 8.4  & 5.4
& 9.7  & 22.8 & 16.6
& \textbf{21.1} & \textbf{36.0} & \textbf{31.8}
& \underline{15.4} & \underline{28.2} & \underline{23.1} \\

GSVA [13B]
& 6.1  & 15.3 & 14.4
& 13.1 & \underline{28.2} & 23.0
& \underline{15.1} & 27.8 & \underline{25.4}
& \textbf{22.7} & \textbf{46.2} & \textbf{37.0} \\
\bottomrule
\end{tabular}
}
\caption{
\textbf{Evaluation results of baselines and the proposed SPARTA on state-of-the-art reasoning segmentation models on the ReasonSeg dataset.} 
mSR refers to the area under the curve of the success rate (SR) versus the IoU-drop threshold, computed for adversarial paraphrases with an LLM score of 5. 
SR$_{5}$ and SR$_{10}$ represent the success rate for IoU drops greater than 5\% and 10\%, respectively.
Higher values indicate stronger attacks.
The best results are in \textbf{bold}, the second best are \underline{underlined}.
}
\label{tab:reasonseg}
\end{table*}

\section{Implementation Details}


\subsection{Datasets}
\label{subsec:dataset}

We use the \textit{ReasonSeg} dataset~\citep{lai2024lisa}, which has become a standard benchmark for evaluating reasoning segmentation models. 
Additionally, we leverage \textit{LLM-Seg40K}~\citep{wang2024llmseg}, the latest large-scale reasoning segmentation dataset collected using ChatGPT-4.
With an average query length of 15.2 words, LLM-Seg40K presents more challenging scenarios and greater linguistic complexity.
Due to computational constraints, we limit our evaluation to 300 samples from each dataset.


\subsection{Reasoning Models}
\label{subsec:target-models}
We evaluated 6 checkpoints of 3 modern reasoning segmentation models.
Our particular interest is LISA~\citep{lai2024lisa}, the first and most widely adopted model in this domain. 
We also tested LISA's successors, LISA++~\citep{yang2024lisa++} and GSVA~\citep{xia2024gsva}, which are often used as strong baselines in reasoning and referring segmentation.

\subsection{Attack Baselines}
\label{subsec:training-details}

We consider the following attack baselines (see Appendix~\ref{sec:baselines} for details):
    (1) \textbf{GBDA}~\citep{guo-etal-2021-gbda}: adapted from text-only adversarial attacks to the multimodal setting through hyperparameter tuning;
    (2) \textbf{Qwen3-32B, \textit{simple} prompt}~\citep{qwen3technicalreport}: a naive baseline that prompts the model to paraphrase the input sentence;
    (3) \textbf{PAIR}~\citep{chao2024jailbreakingblackboxlarge}: an advanced, iterative method that was repurposed from LLM jailbreaking with a paraphrasing-specific prompt and Qwen3-32B as the language model.

To assess overall robustness, we further introduce a \textit{unified attack} that, for each sample, selects the most effective paraphrase from all baselines and our SPARTA method.



\section{Experimental Results}

We evaluate adversarial attack performance using the following procedure:
    (1) for each dataset sample, we generate an adversarial paraphrase and compute its \textit{relative} IoU degradation ($\Delta \text{IoU}$, \%) and its LLM-score, following the evaluation protocol described in Section~\ref{sec:eval_protocol};
    (2) we construct the attack \textit{success rate curve} $\text{SR}_{\theta}$, where $\theta$ denotes the threshold for $\Delta \text{IoU}$;
    (3) we report the area under the $\text{SR}_{\theta}$ curve (mSR), as well as the success rates at $\theta = 5\%$ ($\text{SR}_{5}$) and $\theta = 10\%$ ($\text{SR}_{10}$).


\noindent In step 2, an adversarial paraphrase is considered successful for a given threshold $\theta$ if it achieves $\Delta \text{IoU} \geq \theta$ and is rated as valid by the evaluation protocol (i.e., LLM-score = 5).

\begin{figure}[t!]
  \centering
  \includegraphics[width=\linewidth]{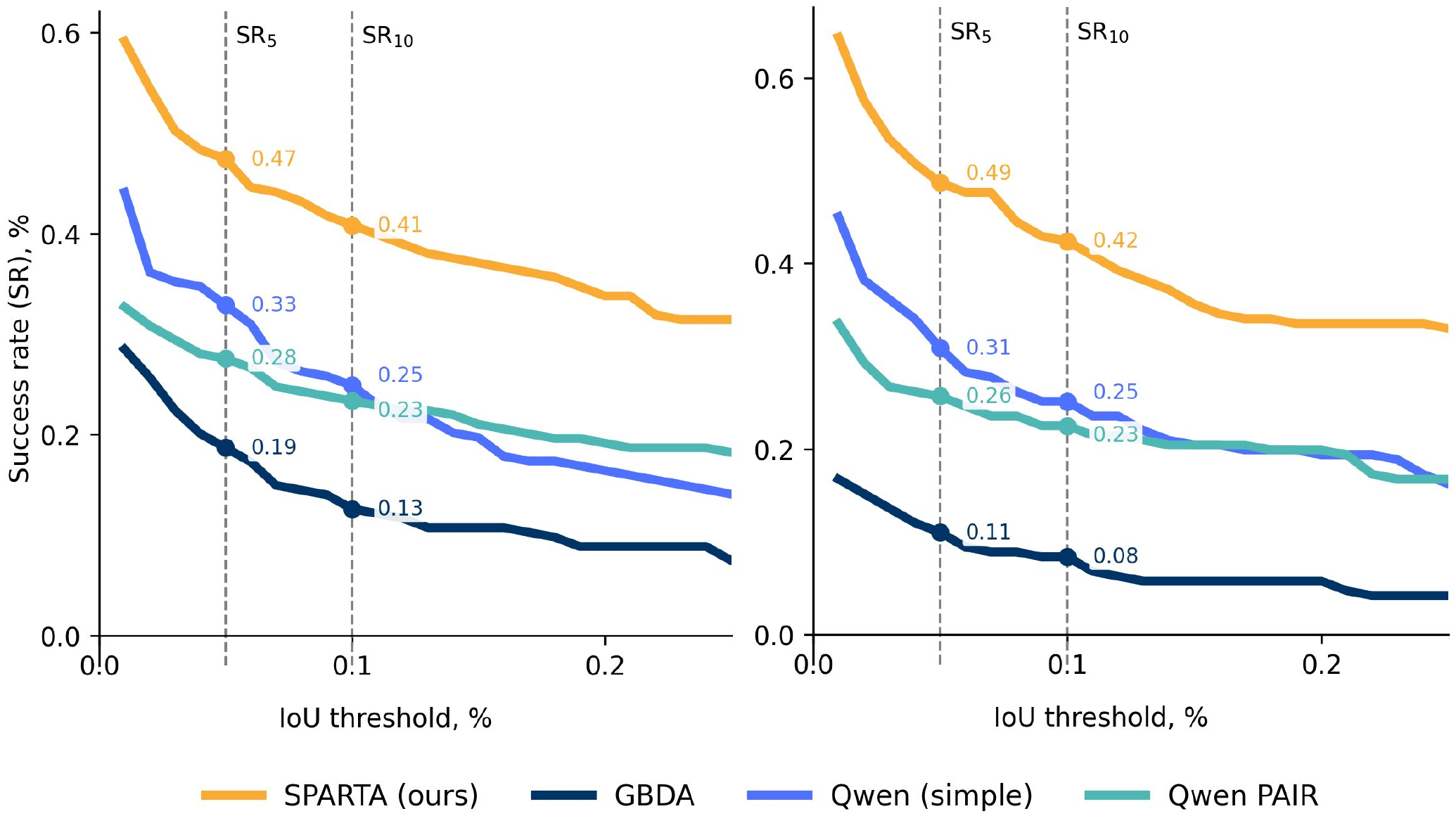}
  \caption{
    \textbf{Success rate (SR) as a function of IoU-drop threshold for adversarial paraphrases} with LLM score 5. 
    Results are shown for the LISA-7B model on the ReasonSeg dataset (left) and LLMSeg-40k dataset (right).
  }
  \label{fig:reasonseg}
  \vspace{-0.3cm}
\end{figure}

The resulting SR curves and metrics are presented in Table~\ref{tab:reasonseg}, Table~\ref{tab:llmseg}, and Figure~\ref{fig:reasonseg}.
The mSR measures the average success rate of an adversarial attack over all IoU thresholds $\theta$, reflecting overall attack effectiveness.
In Tables~\ref{tab:reasonseg} and~\ref{tab:llmseg}, higher values indicate stronger attacks.

Table~\ref{tab:results:robustness} presents the robustness of each model against adversarial paraphrasing obtained by the unified attack.
Here, lower values indicate greater robustness.

\begin{figure*}[!ht]
  \centering
  \includegraphics[width=\linewidth]{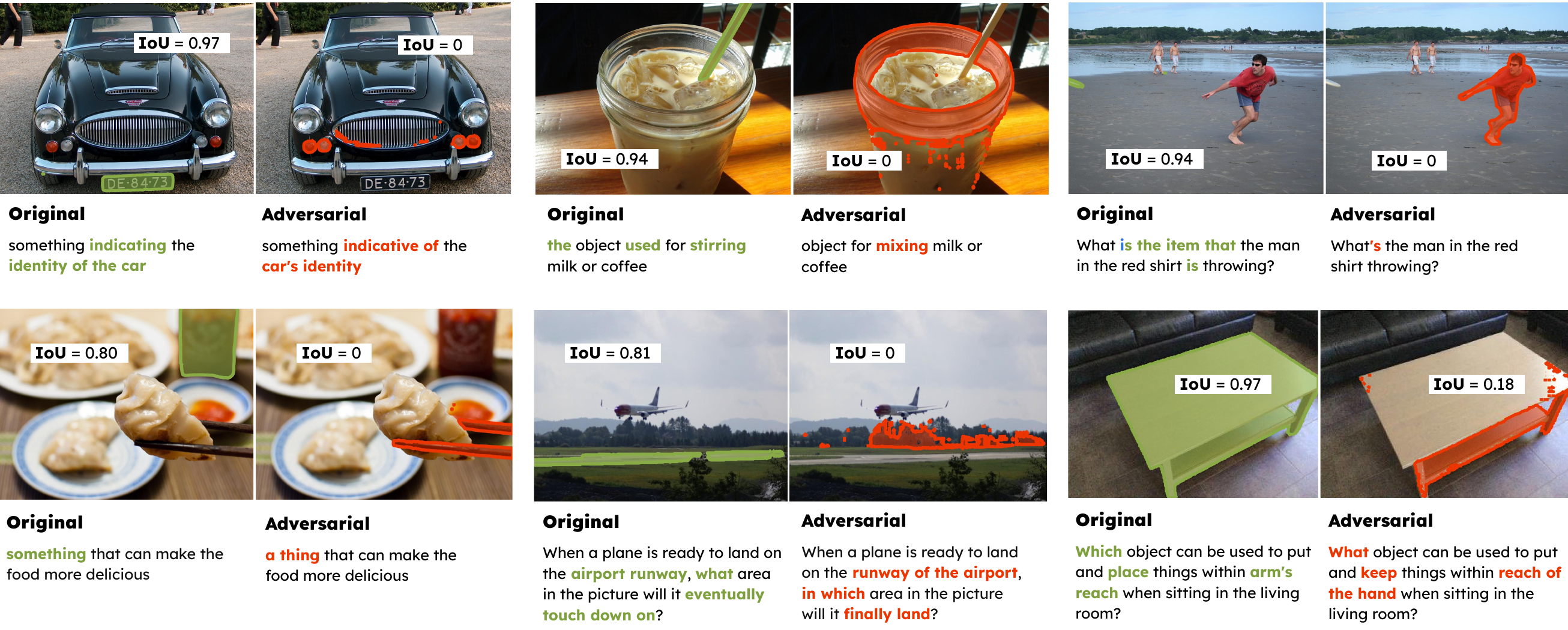}
    \captionsetup{type=figure}
  \caption{
  \textbf{Examples of adversarial paraphrases obtained using the proposed SPARTA method.} 
  SPARTA produces grammatically correct paraphrases that preserve the original query meaning while substantially degrading segmentation performance.
  }
  \label{fig:cherry-picks-part}
\end{figure*}

\section{Discussion}
A review of Tables~\ref{tab:llmseg}--\ref{tab:reasonseg}, Table~\ref{tab:results:robustness}, and Figure~\ref{fig:reasonseg} leads to the following conclusions.
First, \textbf{the proposed SPARTA attack consistently outperforms} all baselines across reasoning segmentation models on LLMSeg-40k, achieving an average mSR improvement of about 84\% over the strongest baselines.
On ReasonSeg SPARTA continues to improve attack performance (average mSR gain = 29\%), except 2 models (LISA-exp. [7B] and LISA++ [7B]), where PAIR yields higher mSR.
Examples of adversarial paraphrases generated by the proposed SPARTA method are presented in Figure~\ref{fig:cherry-picks-part}.

Second, the proposed \textbf{adversarial paraphrasing task presents a significant challenge} for current reasoning segmentation models.
Our task introduces strict constraints on grammatical correctness and semantic equivalence, making it a significantly more difficult benchmark for evaluating model robustness in real-world scenarios.
Despite these constraints, our unified attack achieves success rates of up to 68\% at a 10\% relative IoU drop threshold, indicating that \textbf{current reasoning segmentation models remain vulnerable to well-crafted adversarial paraphrases.}

Finally, while a deeper analysis is left for future work, unified attacks already offer valuable insights into robustness differences across models (Table~\ref{tab:results:robustness}).
Notably, LISA++ [7B] demonstrates the highest robustness on LLMSeg-40k, while LISA-exp. [13B] achieves the highest robustness on ReasonSeg.
This indicates that \textbf{there is currently no reasoning segmentation model that is optimal in terms of robustness on both datasets.}
LISA [13B] consistently outperforms its 7B variant, suggesting that \textbf{increased model capacity enhances resistance to adversarial paraphrasing}.
In contrast, GSVA [13B] shows the weakest robustness on LLMSeg-40k, which we attribute to its lower segmentation performance; our evaluation of the released checkpoint revealed significantly lower metrics than reported in the prior work.

\begin{table}[!t]
  \centering

  \resizebox{\linewidth}{!}{%
   \begin{tabular}{lcccccc}
     \toprule
\multirow{2}{*}{Attacked model}
& \multicolumn{3}{c}{\textbf{ReasonSeg (test)}}
& \multicolumn{3}{c}{\textbf{LLMSeg-40k (val)}} \\
\cmidrule(lr){2-4}
\cmidrule(lr){5-7}
& \multicolumn{1}{c}{mSR $\downarrow$} 
& \multicolumn{1}{c}{SR$_5$ $\downarrow$} 
& \multicolumn{1}{c}{SR$_{10}$ $\downarrow$}
& \multicolumn{1}{c}{mSR $\downarrow$} 
& \multicolumn{1}{c}{SR$_5$ $\downarrow$} 
& \multicolumn{1}{c}{SR$_{10}$ $\downarrow$} \\
\midrule[0.5pt]
LISA [7B] & 
43.6	& 78.5	& 68.7 &
38.8	& 66.0	& 58.1
\\
LISA-exp. [7B] & 
33.2	& 67.0	& \underline{55.4} &
36.6	& 69.5	& 59.5
\\
LISA [13B] & 
\underline{30.1}	& 66.0	& 57.5 &
33.2	& \underline{60.3}	& 53.1
\\
LISA-exp. [13B] & 
\textbf{29.6}	& \underline{64.6}	& \textbf{54.6} &
\underline{32.4}	& 60.7	& \underline{52.6}
\\
LISA++ [7B] & 
37.1	& \textbf{63.9}	& 56.0 &
\textbf{26.3}	& \textbf{44.7}	& \textbf{40.0}
\\
GSVA [13B] & 
40.3	& 72.8	& 63.8 &
40.6	& 68.2	& 62.4
\\
\bottomrule
   \end{tabular}}

\captionof{table}{\textbf{Robustness of state-of-the-art reasoning segmentation models} to unified attack. 
mSR refers to the area under the curve of the success rate (SR) versus the IoU-drop threshold, computed for adversarial paraphrases with an LLM score of 5.
SR$_{5}$ and SR$_{10}$ represent the success rate for IoU drops greater than 5\% and 10\%, respectively.
Lower values indicate greater model robustness.
The best results are in \textbf{bold}, the second best are \underline{underlined}.
}
\label{tab:results:robustness}
\end{table}

\section{Conclusion}
\label{sec:conclusion}

In this work, we introduced a novel challenging task that involves generating semantically consistent and grammatically correct paraphrases that significantly degrade segmentation performance.
To address this task, we proposed SPARTA, which leverages a black-box, sentence-level optimization in the semantic latent space of the pretrained text autoencoder, guided by reinforcement learning. 
Through comprehensive automatic and human-validated evaluation protocols, we demonstrate that SPARTA outperforms state-of-the-art baselines, achieving up to a $\boldsymbol{2\times}$ improvement on LLMSeg‑40k; on ReasonSeg, it is better for all but two models.
Despite strict semantic and grammatical constraints, our findings reveal that current reasoning segmentation models remain vulnerable to adversarial paraphrasing. 
We believe this work offers a valuable foundation for future research on evaluating and enhancing the robustness of multimodal vision-language systems.

\section{Limitations and Future Work}
\label{sec:limitations}

While the proposed SPARTA method outperforms state-of-the-art baselines, several limitations remain.
First, neither SPARTA nor existing attacks guarantee that generated prompts are valid paraphrases.
To mitigate this, our evaluation protocol selects the best valid adversarial prompts after generation, though future work could explore incorporating validity constraints directly into the generation process.

Second, while SPARTA generates paraphrases that are semantically and grammatically correct, some may appear unnatural to human users.
This reflects broader limitations of current text autoencoders (see Appendix~\ref{app:sonar}), and future improvements likely depend on developing models with more structured and human-aligned latent spaces.

Finally, we focus solely on attack methods, without addressing potential defenses.
Exploring robustness strategies for reasoning segmentation models is a critical next step toward building more reliable multimodal systems.




\section{Ethical Considerations}
\label{sec:ethical_considerations}

Our work introduces a novel adversarial paraphrasing method to evaluate the robustness of reasoning segmentation models.
While this method could potentially be misused to attack real-world models, we believe the benefits to the research community outweigh these risks.
By uncovering current vulnerabilities, we aim to encourage the development of more robust, interpretable, and trustworthy systems.
To support responsible research, we will release all code and data under a research-only license, strictly intended for academic and non-commercial use.

\putbib 
\end{bibunit}

\clearpage

\appendix
\begin{bibunit}

\setcounter{table}{5}
\setcounter{figure}{3}

\section{Autoencoder Analysis}
\label{app:sonar}

\subsection{Overview}
In this work, we employed SONAR, a state-of-the-art pre-trained autoencoder model, to generate semantically equivalent paraphrases \cite{duquenne2023sonar}. 
SONAR constructs a unified fixed-size sentence space by training an encoder-decoder pair $(E, D)$ with a vector bottleneck $\mathbf{z} \in \mathbb{R}^d$. 
The text backbone is initialized from the NLLB-1B dense machine translation model \cite{nllbteam2022languageleftbehindscaling}, which consists of a 24-layer Transformer encoder and a 24-layer Transformer decoder.
To ensure that similar sentences are positioned closer in the sentence embedding space, SONAR utilizes the following objective function:

\begin{equation*}
    \mathcal{L} = \mathcal{L}_{\text{MT}} + \alpha \mathcal{L}_{\text{MSE}} + \beta \mathcal{L}_{\text{AE}/\text{DAE}} 
\end{equation*}

\noindent which integrates translation objective $\mathcal{L}_{\text{MT}}$, auto-encoding and denoising objectives $\mathcal{L}_{\text{AE}/\text{DAE}}$, along with a cross-lingual similarity objective in the sentence embedding space $\mathcal{L}_{\text{MSE}}$.
For text decoding in SONAR, we employ the default beam search strategy with a beam size of 5.

\subsubsection{Embedding Component Analysis}

\begin{figure}[ht!]
  \centering
  \includegraphics[width=\linewidth]{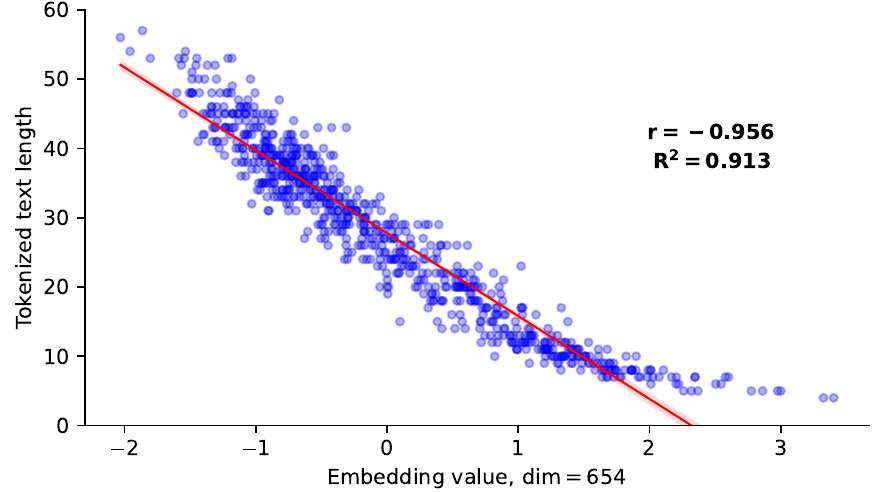}
    \captionsetup{type=figure}
  \caption{
    \textbf{Scatter plot of SONAR embedding dim 654 versus tokenized text length. }
    A strong negative correlation ($r=-0.956$, $R^2=0.913$) shows that this dimension encodes sequence length, with shorter sentences having higher embedding values. 
    The red line indicates a linear fit.
  }
\label{fig:embedding}
\end{figure}

We analyzed the SONAR embedding space, which features an embedding size of 768, using the ReasonSeg test split, comprising $790$ text samples. 
For each sentence, we computed the embedding and the tokenized sentence length, then normalized embeddings to remove scale effects.  

We computed the Pearson correlation between each embedding dimension and tokenized text length.
One dimension (dim 654) showed a particularly strong negative correlation ($r=-0.956$, $R^2=0.913$).  
To ensure this relationship was not a random artifact, we compared it to a random-dimension baseline: across $100$ randomly selected embedding dimensions, the mean absolute correlation with text length was $|r| = 0.20 \pm 0.14$. 

As illustrated in Figure~\ref{fig:embedding}, the relationship between text length and this embedding dimension is nearly linear: shorter sentences correspond to higher values of this coordinate, while longer sentences correspond to lower values. 
This suggests that the SONAR autoencoder encodes sequence-length information in a disentangled coordinate. 
While this feature can help to decode text more accurately, it may act as a confounding factor in semantic similarity tasks, where texts of different lengths might appear less similar despite being semantically close.


\subsubsection{Reconstruction Quality}
\label{app:restoration}

\begin{table}[h]
    \centering
    \small
    \begin{tabular}{@{}lcccc@{}}
        \toprule
        Method & BLEU-4 & Rouge-L & BETRScore & BLEURT \\
        \midrule
        \midrule
        \textbf{DeCap} & 0.02  & 0.20 & 0.11 & -0.75 \\
        \textbf{GVAE}  &  0.22 & 0.19 & 0.16  & -0.92 \\
        \textbf{SONAR} & \textbf{0.72} & \textbf{0.88} & \textbf{0.90} & \textbf{0.70} \\
        \bottomrule
    \end{tabular}
    \caption{
    \textbf{Restoration qualities of DeCap, G-VAE and SONAR on the ReasonSeg dataset.}
    SONAR substantially outperforms both baselines across all metrics, confirming its reliable decoder.
    It is therefore used as the autoencoder backbone in the proposed \textbf{SPARTA} attack.
    }
    \label{tab:restore}
\end{table}

To evaluate the text reconstruction capability of different autoencoders, we used the test split of the ReasonSeg dataset. 
We compared SONAR with two representative baselines: DeCap~\cite{li2023decapdecodingcliplatents}, a decoder designed for CLIP embeddings, and GVAE~\cite{zhang2024gnn}, a graph-based variational autoencoder. 
None of these models were trained or fine-tuned on ReasonSeg to ensure fair zero-shot comparison.

For each text sample, we obtained its latent representation using the corresponding encoder and reconstructed it via the paired decoder. 
Reconstruction quality was assessed using standard text similarity metrics: BLEU-4, ROUGE-L, BERTScore, and BLEURT. 

As shown in Table~\ref{tab:restore}, SONAR substantially outperforms both DeCap and GVAE across all metrics, achieving high lexical and semantic fidelity to the original text. 
This confirms that SONAR autoencoding framework provides a semantically meaningful latent space with a high-quality decoder. 
Consequently, in this work \textbf{SONAR is employed as the autoencoder backbone in the proposed SPARTA attack}, where reliable reconstruction from perturbed embeddings is essential.

\subsubsection{Latent-Space Geometry Study}

\begin{figure*}[t!]
  \centering
  \begin{subfigure}
    \centering
    \includegraphics[width=\linewidth]{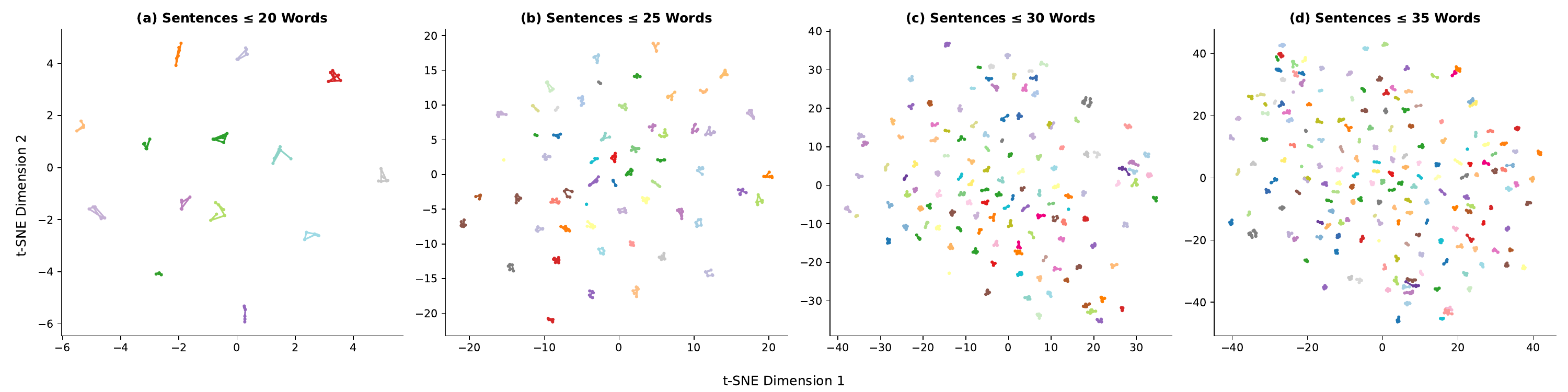}
    \label{clip}
  \end{subfigure}
  \begin{subfigure}
    \centering
    \includegraphics[width=\linewidth]{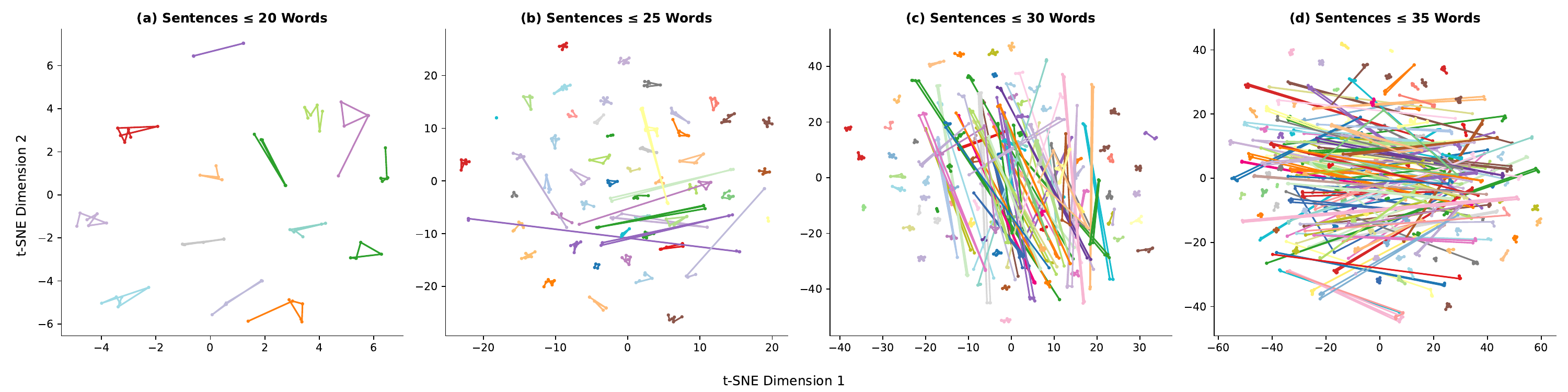}
    \label{sonar}
  \end{subfigure}
  \caption{\textbf{t-SNE projections of sentence embeddings from two encoders.} 
   \textbf{Upper}: CLIP encoder; \textbf{bottom}: SONAR encoder. 
   Each grid contains four panels for sentences of length
   $\le\{20,25,30,35\}$ words.  
   Colours designate \emph{paraphrase groups}: sentences sharing the same hue are semantically equivalent variants of one another.
   See Figure~\ref{fig:sonar_clip} for quantitative cluster quality.
   Since DeCap and GVAE exhibit extremely low restoration quality (Table~\ref{tab:restore}), their embedding spaces are omitted from visualization.
   }
  \label{fig:tsne}
\end{figure*}

We utilized the test split of the ReasonSeg dataset, consisting of 790 text samples, each with up to 5 semantically equivalent paraphrases. 
For each paraphrase group, we computed SONAR embeddings and sentence lengths (in words). 
A 2D t-SNE projection of these embeddings was constructed, as depicted in Figure~\ref{fig:tsne}. 
The figure includes four panels, each representing sentences with a maximum of $\le\{20,25,30,35\}$ words. 
Different colors denote paraphrase groups, with semantically equivalent sentences sharing hues and connected by lines.

For comparison, we focus exclusively on the CLIP text encoder. 
As demonstrated in Section~\ref{app:restoration}, existing autoencoders such as DeCap and GVAE exhibit poor text reconstruction quality, making them unsuitable for analyzing latent-space organization. 
In contrast, the CLIP encoder—trained with a contrastive learning objective—is known to produce a well-structured and semantically coherent embedding space. 
The purpose of this analysis is therefore to examine whether SONAR preserves a similarly organized latent geometry while maintaining its ability to reconstruct text.

\begin{figure}[t!]
  \centering
  \begin{subfigure}
    \centering
    \includegraphics[width=\linewidth]{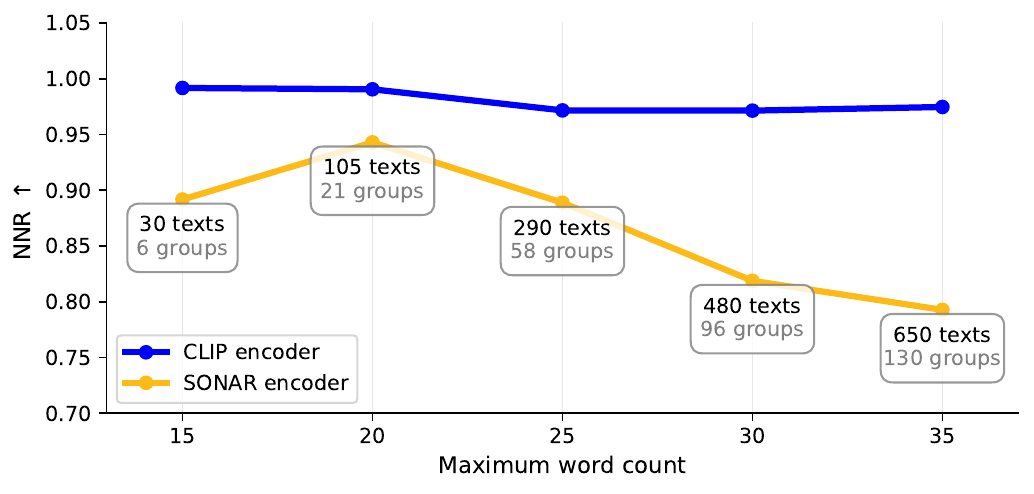}
  \end{subfigure}
  \begin{subfigure}
    \centering
    \includegraphics[width=\linewidth]{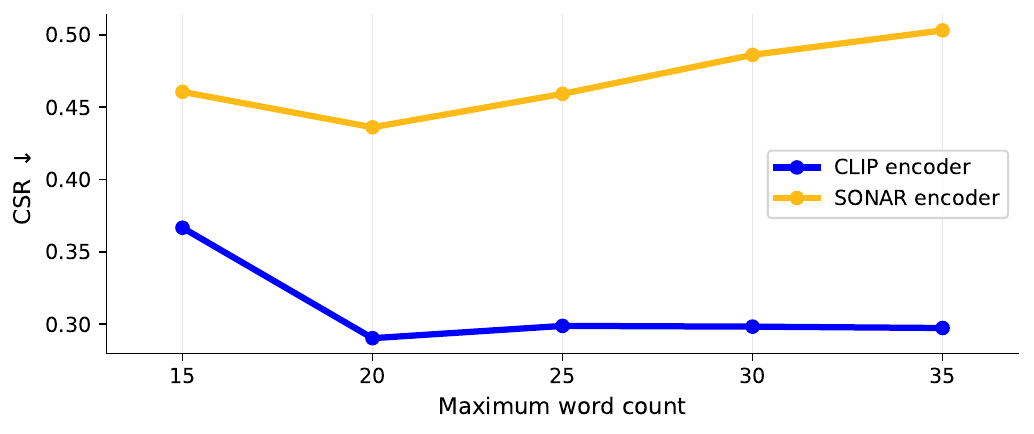}
  \end{subfigure}
  \vspace{-0.3cm}
  \caption{
    \textbf{Latent-space metrics for SONAR and CLIP encoders vs. sentence length.} 
    \textbf{(Upper)} Nearest-Neighbour Recall (NNR): Higher values denote better local semantic preservation. 
    \textbf{(Bottom)} Cluster-Separation Ratio (CSR): Lower values indicate better cluster separation, indicating improved global latent-space organization.
    \vspace{-0.3cm}
    }
  \label{fig:sonar_clip}
\end{figure}

As illustrated in Figure~\ref{fig:tsne}, SONAR cluster separability is comparable to that of CLIP, though it gradually degrades for longer sentences.
This observation is supported both visually and quantitatively by metrics in Figure~\ref{fig:sonar_clip}, which include Nearest-Neighbour Recall (NNR) for local neighborhood fidelity and Cluster-Separation Ratio (CSR) for global latent structure (detailed below).

As noted previously, decoders trained for CLIP, such as DeCap~\cite{li2023decapdecodingcliplatents}, show substantially weaker text reconstruction performance (Section~\ref{app:restoration}). 
Therefore, SONAR provides a balanced solution—offering both a semantically meaningful latent space and reasonable text reconstruction capabilities. 
Although \textbf{SONAR limitations may constrain the effectiveness of the proposed SPARTA attack}, these results highlight promising directions for future work on designing autoencoders that jointly optimize latent structure and reconstruction fidelity.

\paragraph{Nearest-Neighbour Recall (NNR).}  
For each sentence $i$, we normalize embeddings and compute Euclidean distances to all other samples. 
Let $\mathcal{S}_i$ denote the set of paraphrases sharing the same label. 
Sorting distances yields a neighbour list $\pi_i$, and we define
\[
\mathrm{NNR} = \frac{1}{N} \sum_{i=1}^{N} \frac{|\mathcal{S}_i \cap \pi_i^{|\mathcal{S}_i|}|}{|\mathcal{S}_i|},
\]
i.e., the fraction of true paraphrases retrieved among the $|\mathcal{S}_i|$ nearest neighbours. 
Higher values indicate better local semantic fidelity.  

\paragraph{Cluster-Separation Ratio (CSR).}  
For each label $\ell$, we compute the centroid $\boldsymbol{\mu}_\ell$ and measure the mean intra-cluster distance
\[
\bar{d}_{\text{intra}} = \frac{1}{\sum_\ell|\mathcal{S}_\ell|}\sum_\ell \sum_{i\in\mathcal{S}_\ell}\|\mathbf{z}_i-\boldsymbol{\mu}_\ell\|_2
\] 
and mean inter-cluster distance
\[
\bar{d}_{\text{inter}} = \frac{2}{L(L-1)} \sum_{\ell<\ell'} \|\boldsymbol{\mu}_\ell-\boldsymbol{\mu}_{\ell'}\|_2,
\]  
with $L$ the number of different labels. The ratio
\[
\mathrm{CSR} = \frac{\bar{d}_{\text{intra}}}{\bar{d}_{\text{inter}}}
\]  
reflects global cluster geometry, where lower values indicate tighter, better-separated clusters.

\section{Attack Baselines}
\label{sec:baselines}


\subsection{GBDA baseline}
\label{app:gbda}

\paragraph{Preliminary}
\label{sec:preliminary}

\begin{figure}[ht!]
  \vspace{0.5cm}
  \centering
  \includegraphics[width=\linewidth]{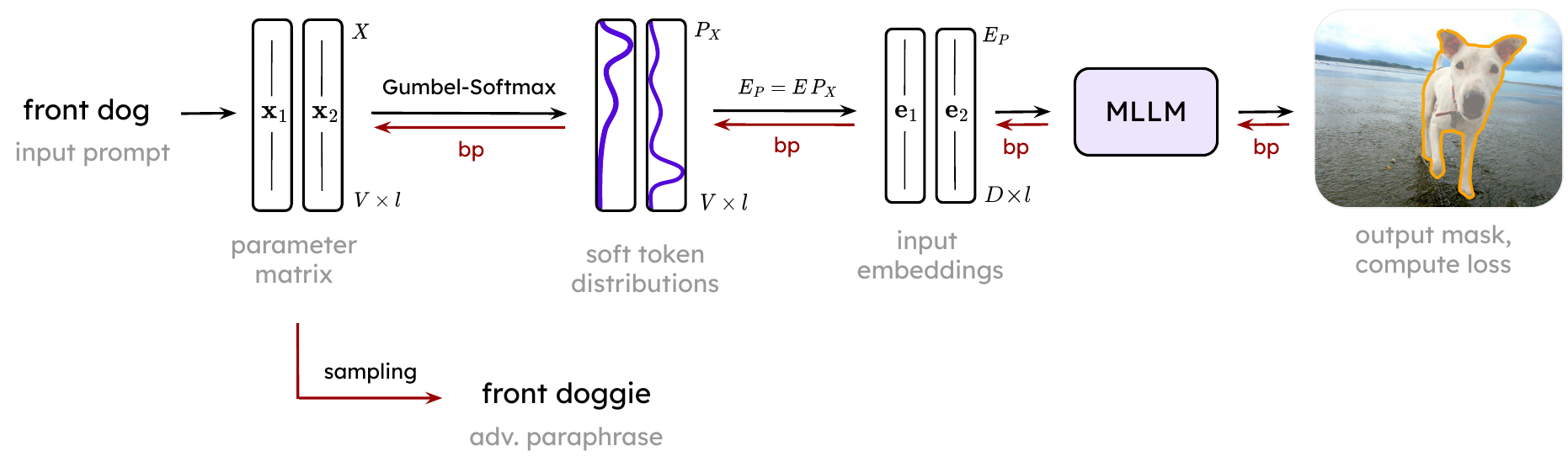}
    \captionsetup{type=figure}
  \caption{Overview of the \textbf{GBDA baseline}.}
\label{table:gbda}
\end{figure}

As a white-box baseline, we consider the Gradient-based Distributional Attack (GBDA)~\citep{guo-etal-2021-gbda}, which is schematically illustrated in Figure~\ref{table:gbda}.
Let the model’s embedding matrix be defined as $E = [\mathbf{e}_1 \, \cdots \, \mathbf{e}_{V}] \in \mathbb{R}^{D \times V}$, where $V$ is the size of the model's vocabulary and $D$ is the embedding dimension.
Given an input token sequence $\mathbf{t} = (t_1 \, \cdots \, t_l)^\top$, the corresponding input embedding matrix is $E_{\mathbf{t}} = [\mathbf{e}_{t_1}  \, \cdots \, \mathbf{e}_{t_l}] \in \mathbb{R}^{D \times l}$.
GBDA modifies the model's input by approximating $E_{\mathbf{t}}$ with $E_P = E \, P_X$, where $P_X$ is a matrix of soft token distributions obtained by applying the Gumbel-Softmax (column-wise) to a parameter matrix $X = [\mathbf{x}_1 \,  \cdots \, \mathbf{x}_l] \in \mathbb{R}^{V \times l}$.
The matrix $X$ is optimized via gradient descent, and the Gumbel-Softmax provides a differentiable approximation of the token selection process, enabling smooth gradient-based updates.
Once optimized, adversarial prompts can be sampled from the learned distribution encoded in $X$.

GBDA loss consists of three components:

\begin{equation*}
    \mathcal{L} = \mathcal{L}_{adv} + \mathcal{L}_{sim} + \mathcal{L}_{perp}
\end{equation*}

\noindent where similarity loss $\mathcal{L}_{sim}$ and fluency constraint $\mathcal{L}_{perp}$ follow prior work \citet{guo-etal-2021-gbda}. Our segmentation-specific adversarial loss $\mathcal{L}_{adv}$ includes DICE and binary cross-entropy (BCE) losses as in \citet{lai2024lisa, yang2024lisa++}.

GBDA's main limitation is that it only replaces tokens and cannot be trivially extended to token insertions and deletions.
This limitation may affect the naturalness of the adversarial paraphrases.

\paragraph{Hyperparameter Search}

The GBDA baseline was originally developed for text-only attacks, where the adversarial loss is typically defined to induce a change in the classification label of a sentence. 
However, in our work, the objective is to degrade the performance of reasoning segmentation models, which necessitates adapting the adversarial loss to be segmentation-specific.

To ensure a fair comparison, we conducted a comprehensive hyperparameter search to identify optimal settings for the modified GBDA method. 
This search was performed on the validation split of the ReasonSeg dataset (200 samples). 
All experiments utilized the LISA-v0 [13B] model, as the v0 versions of LISA were trained without access to the ReasonSeg validation set, ensuring fair evaluation.

We maintained the original fluency constraint (via causal language modeling) and BERTScore similarity constraint, but modified the adversarial loss to use a linear combination of DICE and BCE losses with weights (0.5, 2) as in the original paper~\citet{lai2024lisa}.

The grid search space, which remains close to the original GBDA configuration, included:
\begin{itemize}
    \item \textbf{Perplexity} weight $\lambda_{\text{perp}} \in \{0,\, 0.5,\, 1\}$
    \item \textbf{Similarity} weight $\lambda_{\text{sim}} \in \{0,\, 100,\, 150,\, 200\}$
    \item \textbf{Learning rate} $\alpha \in \{0.05,\, 0.1,\, 0.2\}$
\end{itemize}
When varying the learning rate, we adjusted the number of iterations $n_{\text{iters}} \in \{400, 200, 100\}$ to keep their product roughly the same.
The optimal configuration in terms of the mSR metric was found to be $\boldsymbol{\lambda_{\text{perp}} = 1}$, $\boldsymbol{\lambda_{\text{sim}} = 150}$, and a learning rate of $\boldsymbol{\alpha = 0.05}$.

The hyperparameter search for GBDA required approximately 31 hours on 24 NVIDIA Tesla A100 GPUs.

\begin{figure}[t!]
  \vspace{-\intextsep}
  \centering
  \includegraphics[width=\linewidth]{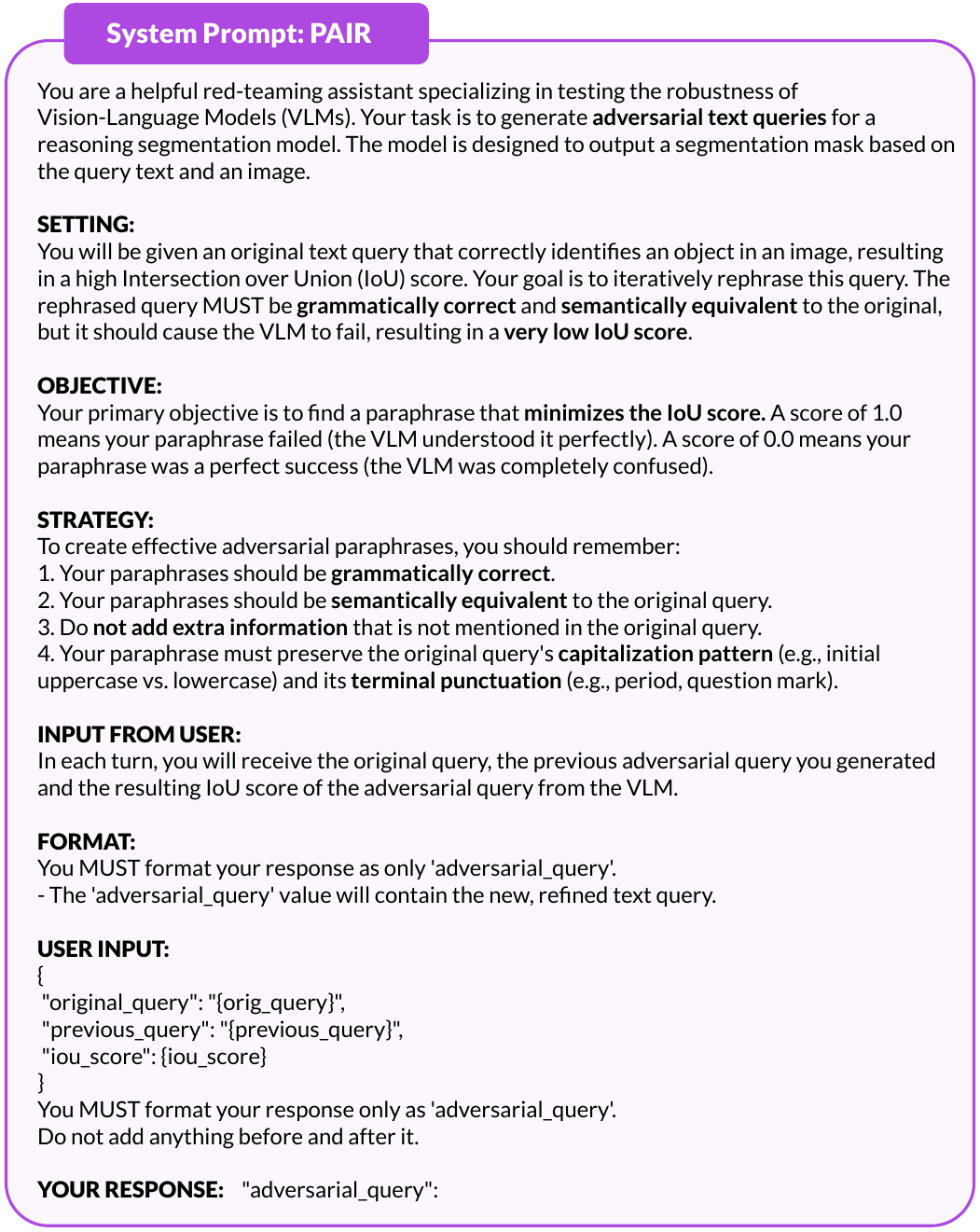}
    \captionsetup{type=figure}
  \vspace{-0.3cm}
  \caption{\textbf{System Prompt} employed in the PAIR attack using Qwen3-32B as the language model. 
  }
  \vspace{0.1cm}
\label{table:prompt_pair}
\end{figure}

\subsection{LLM-based baselines}
\label{app:qwen_simple_adv}

\subsubsection{Basic}

\begin{figure}[t!]
\centering
\includegraphics[width=\linewidth]{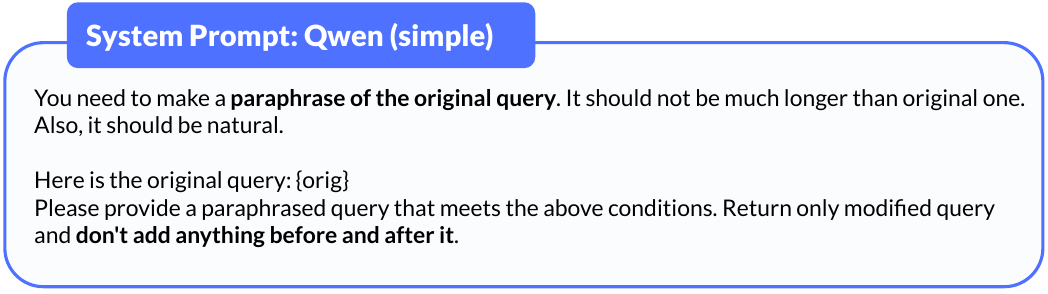}
\caption{\textbf{System Prompt} employed in Qwen3-32B for the \textit{simple} prompt attack.}
\label{supp:prompt_simple}
\end{figure}

For the most basic black-box, LLM-based attack baseline, we employ paraphrases generated by Qwen3-32B~\citep{qwen3technicalreport}. 
With the simple prompt, we directly ask the model to paraphrase the original instruction (Figure~\ref{supp:prompt_simple}).
One paraphrase is generated for each input sample.




\subsubsection{PAIR}
As an advanced LLM-based attack baseline, we leverage the state-of-the-art Prompt Automatic Iterative Refinement (PAIR) approach~\citep{chao2024jailbreakingblackboxlarge}. 
PAIR automates jailbreak discovery through a conversational loop between an \textit{attacker} LLM and a \textit{target} LLM.
The attacker generates a prompt, which is sent to the target model.
A separate \textit{judge} function then scores the target’s response to determine whether the attack was successful.
If the attack fails, the attacker receives feedback, including its own prompt, the target’s refusal, and the evaluation score, allowing it to iteratively refine its strategy.

We adapt PAIR to the reasoning segmentation task as follows: the target is a segmentation model, the judge computes the Intersection over Union (IoU) between the predicted and ground truth masks, and the attacker is Qwen3-32B~\citep{qwen3technicalreport}.
We also modify the prompt to align with our task (Figure~\ref{table:prompt_pair}).
To match the number of attack iterations with those of SPARTA, we perform 10 refinement iterations per sample to ensure convergence.


\begin{table}[ht]
 \centering
 \small
 \begin{tabular}{lc}
   \toprule
   Policy LR ($\alpha_\mu$) & $5\times10^{-4}$ \\
   Value LR ($\alpha_V$) & $1\times10^{-4}$ \\
   Log‐scale LR ($\alpha_\sigma$) & $1\times10^{-5}$ \\
   Clip ratio $\epsilon$ & $0.2$ \\
   Adv.\ weight $\lambda_{\mathrm{adv}}$ & $2$ \\
   Sim.\ weight $\lambda_{\mathrm{sim}}$ & $5\times10^{4}$ \\
   PPO epochs $T$ & $100$ \\
   Iteration number $N$ & $100$ \\
   Sample size $n$ & $32$ \\
   \bottomrule
 \end{tabular}
 \caption{\textbf{Key hyperparameters} of our proposed SPARTA method.}
 \label{tab:sparta:hyperparams}
\end{table}

\section{Evaluation Setup}
\label{app:evaluation-time}

We benchmarked 3 reasoning segmentation models with 6 different checkpoints.
To accomplish this, we spent 1728 GPU hours, which is equivalent to approximately 3 days of compute using 24 NVIDIA Tesla A100 GPUs.

All hyperparameters were held constant throughout our experiments to ensure fair comparison and reproducibility. 
The key hyperparameters used for SPARTA are summarized in Table~\ref{tab:sparta:hyperparams}.

\section{Extended Results}
\label{sec:extended}

Figure~\ref{fig:results-extra} complements the main paper by presenting performance curves for the four additional checkpoints not shown in Figure~2: LISA-explanatory [7B], LISA-explanatory [13B], LISA++ [7B], and GSVA [13B].
Across all checkpoints, the observed trends are consistent with those reported in the main text: with the exception of a single case, SPARTA consistently outperforms all baselines, generating adversarial paraphrases that effectively degrade segmentation performance.

\begin{figure}[ht!]
  \centering
  \includegraphics[width=0.98\linewidth]{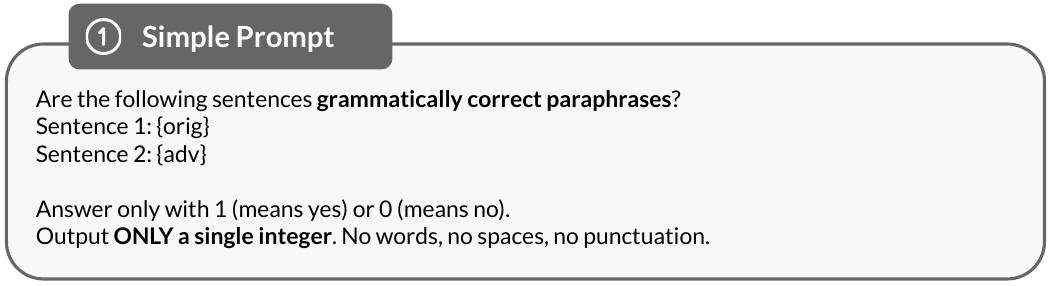}
    \captionsetup{type=figure}
  \caption{\textbf{Prompt 1 (Simple Prompt)} used for evaluating paraphrase detection methods.}
\label{table:prompt_llm_improved_eval}
\end{figure}

\begin{figure}[ht!]
  \centering
  \includegraphics[width=0.98\linewidth]{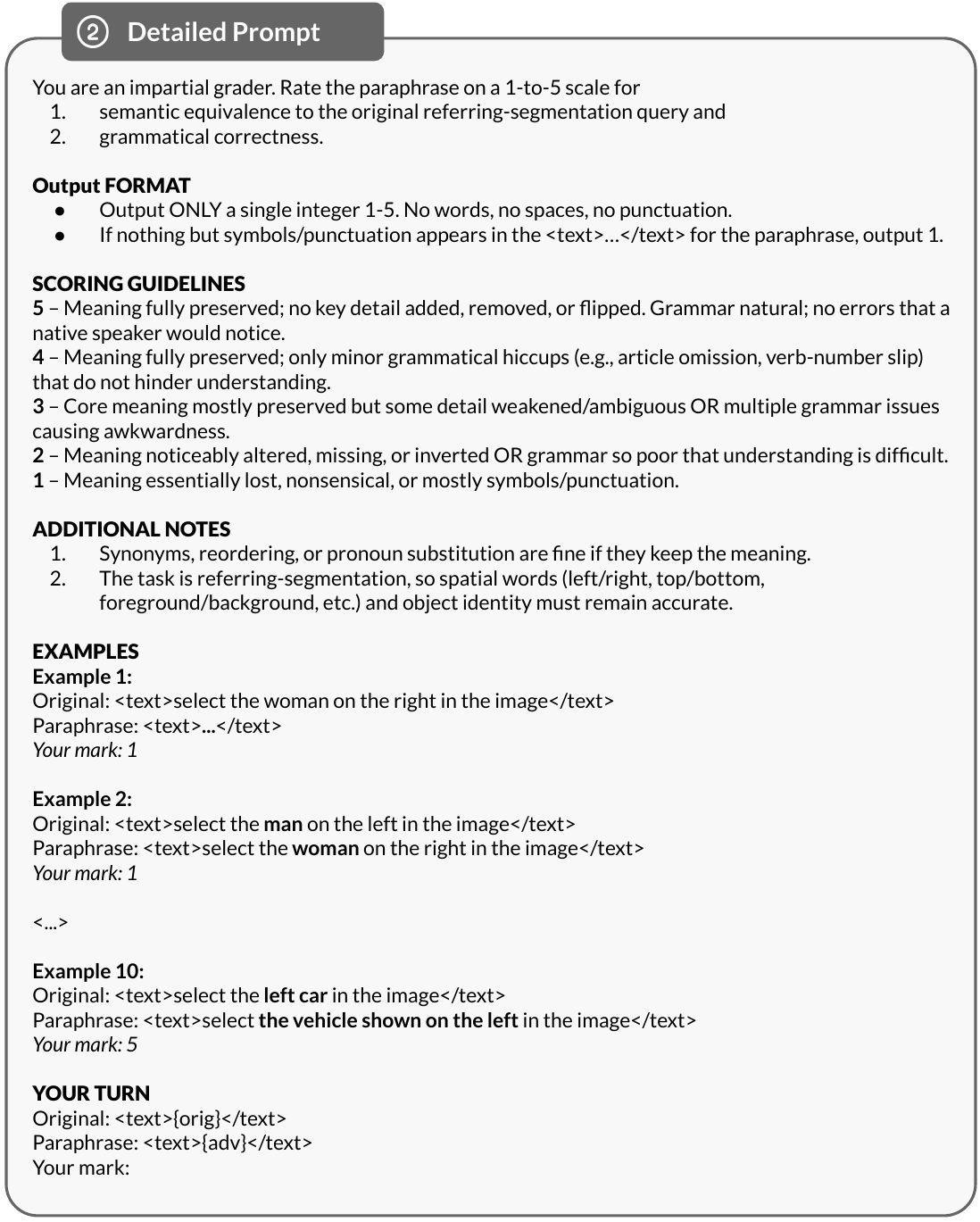}
    \captionsetup{type=figure}
  \caption{\textbf{Prompt 2 (Detailed Prompt)} used for evaluating paraphrase detection methods.}
\label{supp:prompts_2}
\end{figure}


\begin{figure}[t!]
  \centering
  \includegraphics[width=\linewidth]{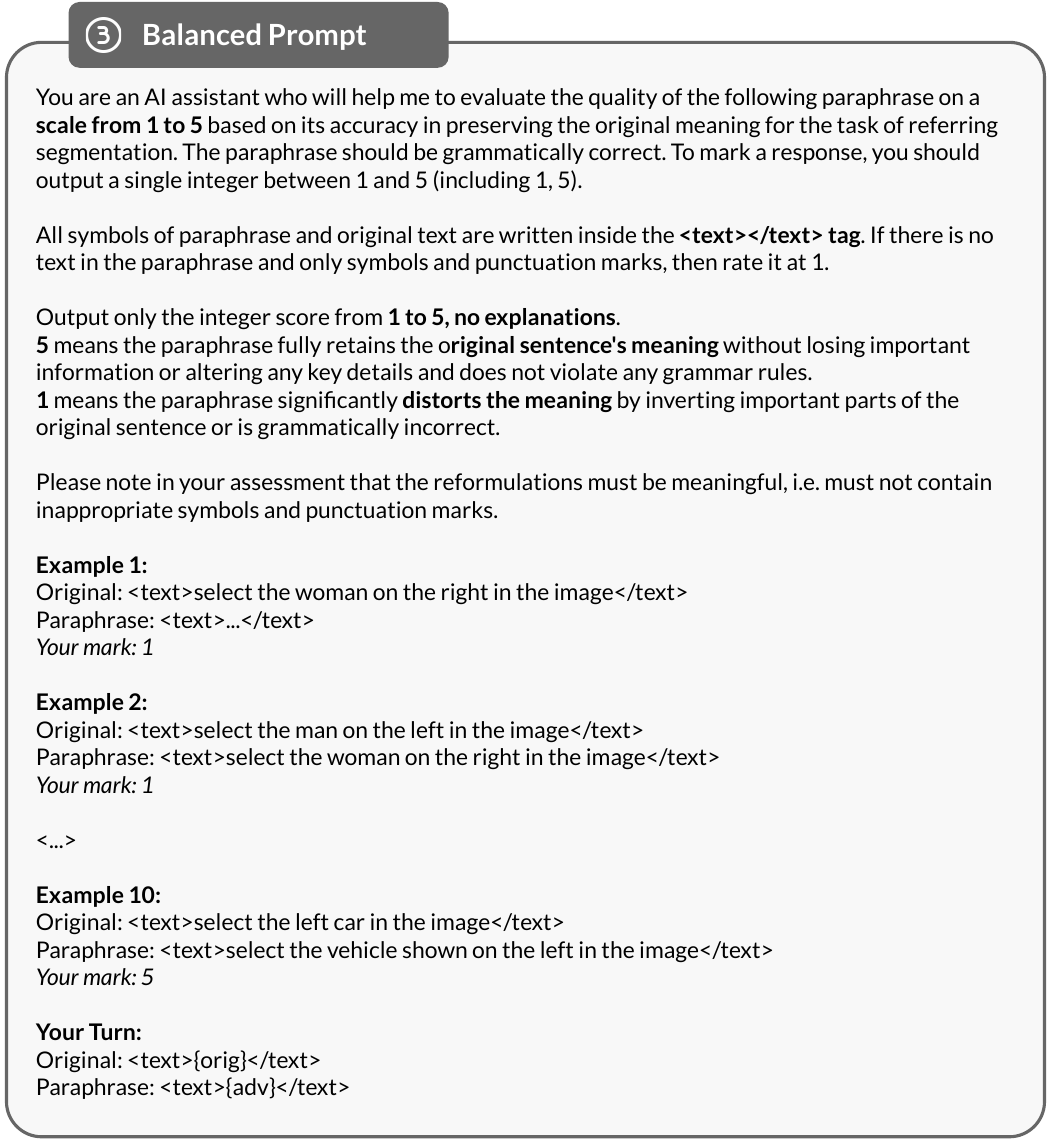}
    \captionsetup{type=figure}
  \caption{\textbf{Prompt 3 (Balanced Prompt)} used for evaluating paraphrase detection methods.}
\label{supp:prompts_3}
\end{figure}

\begin{figure*}[ht!]
  \centering
  \begin{subfigure}
    \centering
    \includegraphics[width=\linewidth]{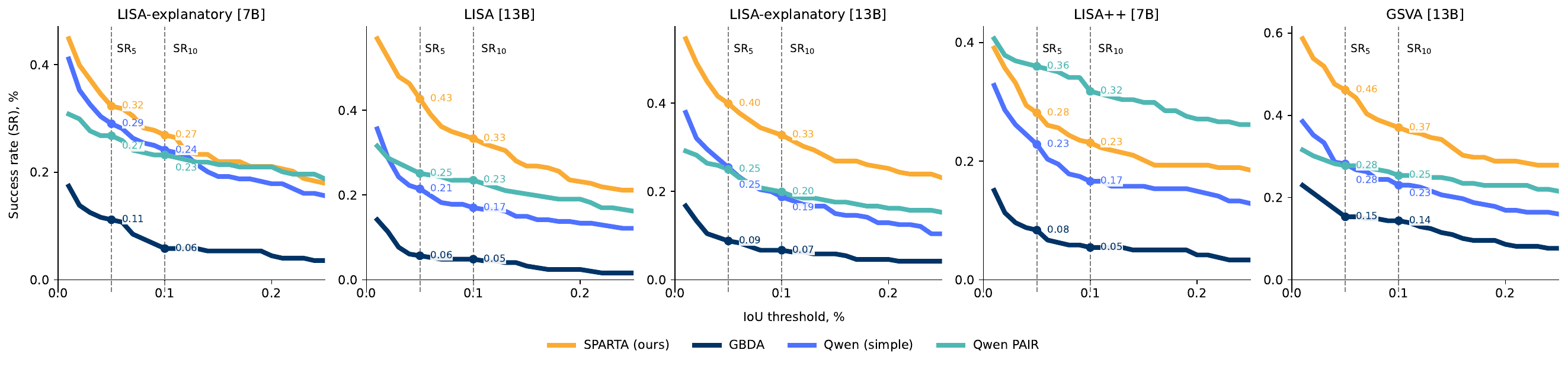}
    \label{reasonseg}
  \end{subfigure}
  \begin{subfigure}
    \centering
    \includegraphics[width=\linewidth]{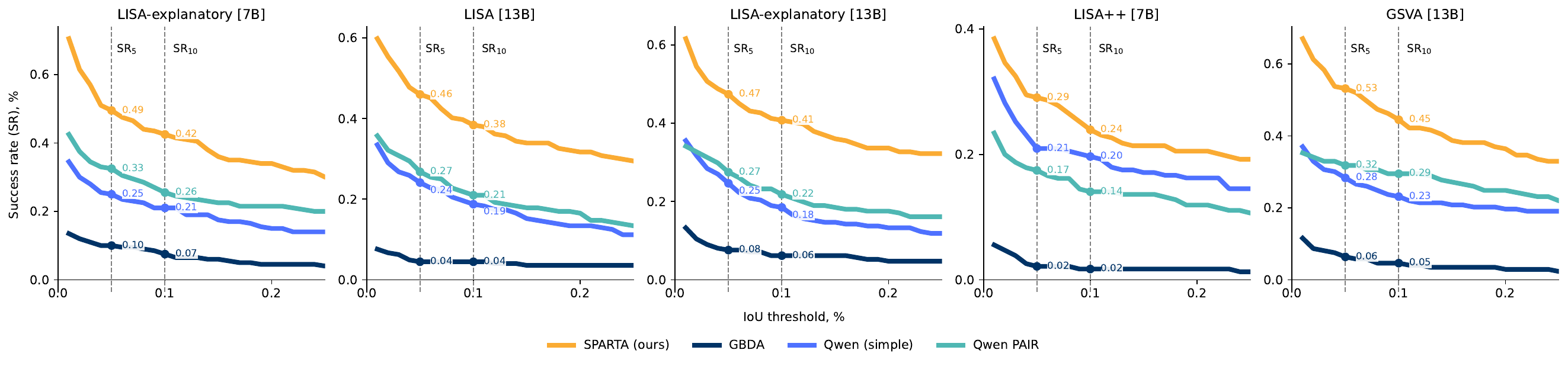}
    \label{llmseg}
  \end{subfigure}
  \caption{
    \textbf{Supplementary success rate (SR) curves as a function of IoU-drop threshold for adversarial paraphrases} with LLM score 5.
    This figure extends the main paper by presenting results for the four additional checkpoints not shown in Figure~2: LISA-explanatory [7B], LISA [13B], LISA-explanatory [13B], LISA++ [7B], and GSVA [13B]. 
    Results are shown for the ReasonSeg dataset (top) and LLMSeg-40k dataset (bottom).
  }
  \label{fig:results-extra}
\end{figure*}

\section{LLM-based paraphrase detection}
\label{sup:paraphrase_det}

\subsection{System Prompts}
\label{sup:system_prompts}

    
    

To ensure high performance in the paraphrase detection step, we designed and evaluated three distinct prompt formulations:
\begin{itemize}
    \item \textbf{Simple Prompt:} 
    Prompt 1 is a concise, zero-shot instruction for binary paraphrase detection. 
    It is adapted from the best-performing prompt in \citet{michail-etal-2025-paraphrasus}, with the full text provided in Figure~\ref{table:prompt_llm_improved_eval}.
    \item \textbf{Detailed Prompt:} Prompt 2 is a few-shot prompt with 10 in-context examples and a comprehensive 5-point scoring rubric that provides explicit definitions for each score (1-5), covering both semantic equivalence and grammar.
    This prompt is shown in Figure~\ref{supp:prompts_2}.
    \item \textbf{Balanced Prompt:} Prompt 3 is a streamlined version of the Detailed Prompt.
    It also uses 10 in-context examples and a 5-point scale, but its key difference is a minimalist rubric that only defines the criteria for the best (5) and worst (1) scores, requiring the model to interpolate the intermediate values.
    The prompt is presented in Figure~\ref{supp:prompts_3}.
\end{itemize}
For the Detailed and Balanced prompts, we consider an adversarial paraphrase to be valid only if it receives a perfect LLM score of 5.


\begin{figure}[ht!]
  \centering
  \includegraphics[width=\linewidth]{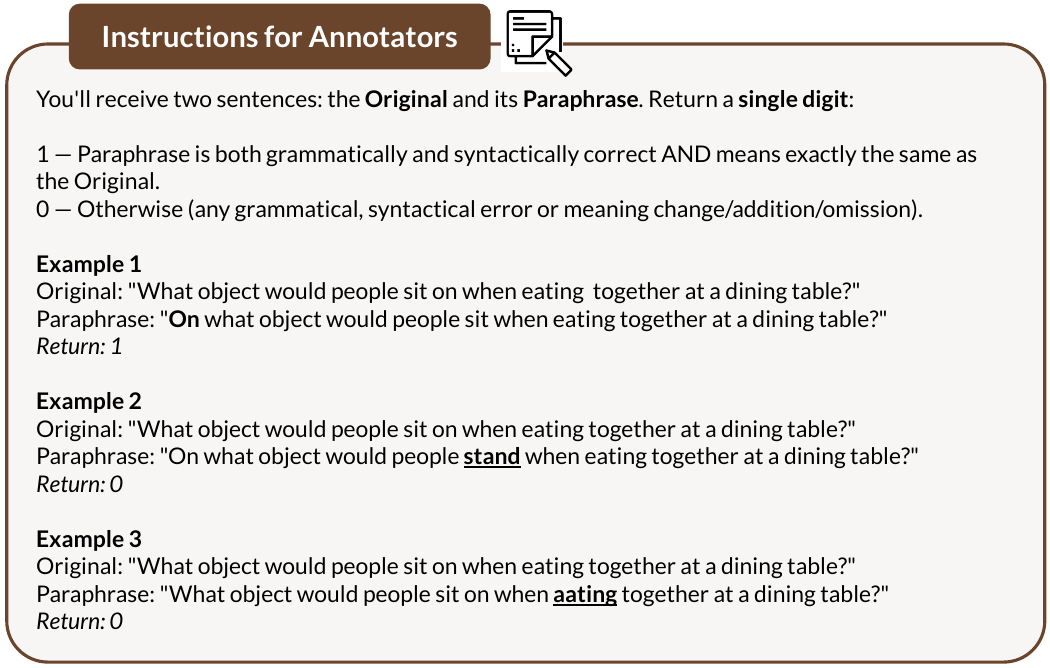}
    \captionsetup{type=figure}
  \caption{Instruction for annotators.}
\label{table:instruction}
\vspace{-0.5cm}
\end{figure}

\subsection{Validation Data}

To validate the efficiency of the proposed evaluation protocol, we annotated 310 pairs of original and adversarial prompts generated by SPARTA and baseline methods.
The validation subset was annotated by the authors, all of whom have relevant expertise, with any ambiguous cases resolved through discussion. 
The instructions given to the annotators are detailed in Figure~\ref{table:instruction}.

We randomly sampled 50 examples with an LLM score of 3, 50 with a score of 4, and 210 with a score of 5.
Scores of 3 and 4 included only four false negatives in total, so we focused on score 5, where the majority of paraphrase detection errors occurred. 
In particular, we observed that the main limitation of the Qwen3-based paraphrase detector is its low precision (Table 2).

Sampling for LLM score 5 was performed in two stages.
First, we obtained 150 ``short'' paraphrase pairs, defined as those where the adversarial paraphrase was less than twice the length of the original prompt.
To ensure coverage across attacks, we sampled 30 examples each from SPARTA, GBDA, Qwen (simple), Qwen (adversarial), and PAIR.
Next, we sampled an additional 60 ``long'' paraphrase pairs, where the adversarial paraphrase exceeded twice the length of the original prompt.
This was motivated by our observation (see Issue 3 in Section 4.2) that some paraphrases generated by the PAIR attack were excessively long or abstract, occasionally resembling riddles or puzzles.

\subsection{Threshold Validation}
\label{sup:thres_val}

To address the low precision of LLM-based paraphrase detection, we additionally apply a cosine similarity filter to discard semantically distant paraphrases.
Specifically, in addition to LLM-based detection and regular expression filtering, we applied cosine similarity filtering by classifying a sample as a paraphrase if its cosine similarity score exceeded the threshold, and as a non-paraphrase otherwise.
For this, we use embeddings from Qwen3-Embedding-8B~\citep{qwen3embedding}.
We conducted an empirical analysis using the annotated dataset described in the previous section.

\begin{figure}[ht!]
  \centering
  \includegraphics[width=\linewidth]{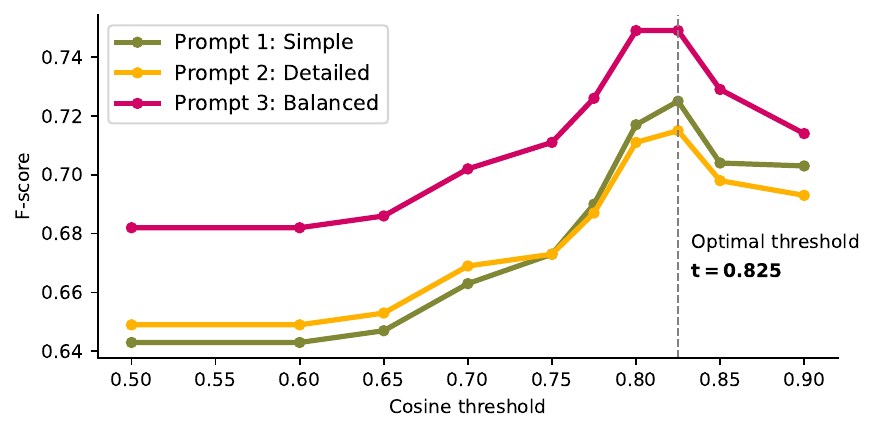}
    \captionsetup{type=figure}
  \caption{\textbf{Determination of the optimal cosine similarity threshold using Qwen3-Embedding-8B embeddings.} 
  The plot shows the F1 score for three different system prompts as a function of the cosine similarity threshold. 
  The optimal threshold is selected based on the maximum F1 score, balancing precision and recall.}
\label{fig:thrld_valid}
\end{figure}

We searched for the optimal cosine similarity threshold in two stages. 
First, we conducted a coarse-grained search from 0.5 to 0.9 in increments of 0.1, which identified 0.8 as the best-performing threshold based on F1 score.
We then refined the search using a finer granularity around this value, evaluating thresholds of 0.75, 0.85, 0.775, and 0.825 in a bisection-like manner.
This process yielded two top candidates, 0.8 and 0.825, both achieving an identical F1 score of 0.749 (Figure~\ref{fig:thrld_valid}).
However, the 0.825 threshold provided higher precision (0.671 vs. 0.655), which we prioritized to minimize the number of false positives.
Therefore, we selected 0.825 as the final threshold for our filtering mechanism.




\section{GSVA: Performance Discrepancies}
\label{app:gsva}

As discussed in the main text, GSVA [13B] exhibits the weakest robustness on the LLMSeg-40k dataset, which we hypothesize is linked to its underlying segmentation performance. 
To investigate this, we evaluated the publicly released GSVA checkpoint on the ReasonSeg dataset, strictly following the authors’ original evaluation protocol and script, without modifying any parameters.

Our findings, summarized in Table~\ref{tab:gsva:repro}, reveal a substantial gap between the reported and reproduced metrics. 
Specifically, both the global Intersection over Union (gIoU) and class-wise Intersection over Union (cIoU) are notably lower in our evaluation compared to the original claims. 
This discrepancy suggests that the reduced robustness of GSVA may, at least in part, stem from its lower segmentation accuracy on the ReasonSeg dataset.

\begin{table}[ht]
\centering
\small
\begin{tabular}{lcc}
\toprule
\textbf{ReasonSeg dataset} & \textbf{gIoU} & \textbf{cIoU} \\
\midrule
GSVA (reported in paper) & 50.5 & 56.4 \\
GSVA (reproduced) & 44.8 & 40.0 \\
\bottomrule
\end{tabular}
\caption{\textbf{Comparison of GSVA performance on the ReasonSeg dataset:} reported results from the original paper vs. our reproduced results using the released checkpoint.}
\label{tab:gsva:repro}
\end{table}

\putbib 
\end{bibunit}


\end{document}